\documentclass[10pt,twocolumn,letterpaper]{article}

\usepackage[pagenumbers]{cvpr} 
\usepackage[accsupp]{axessibility}  

\usepackage[utf8]{inputenc} 
\usepackage[T1]{fontenc}    
\usepackage{url}            
\usepackage{booktabs}       
\usepackage{amsfonts}       
\usepackage{nicefrac}       
\usepackage{microtype}      
\usepackage{xcolor}
\usepackage{soul}
\usepackage{subcaption}
\makeatletter
\@namedef{ver@everyshi.sty}{}
\makeatother
\usepackage{tikz}
\usepackage{pgfplots}
\usepackage{wrapfig}
\usepackage{graphicx}
\usepackage{float}
\usepackage{multirow}

\usepackage{listings,newtxtt}

\definecolor{cvprblue}{rgb}{0.21,0.49,0.74}
\usepackage[pagebackref,breaklinks,colorlinks,allcolors=cvprblue]{hyperref}

\def\paperID{15935} 
\def\confName{CVPR}
\def\confYear{2025}

\definecolor{deepblue}{rgb}{0,0,0.6}
\definecolor{deepred}{rgb}{0.6,0,0}
\definecolor{deepgreen}{rgb}{0,0.5,0}
\lstdefinestyle{python}{
language=Python,
basicstyle=\ttfamily\scriptsize,
commentstyle=\color{deepred},
otherkeywords={},             
keywordstyle=\color{deepgreen},
emph={},          
emphstyle=\color{deepblue},    
stringstyle=\color{deepred},
showstringspaces=false            %
}

\newcommand{\textlabel}[1]{{\scriptsize #1}}

\newcommand{\rawdata}[1]{\csname data-#1\endcsname}

\newcommand{\defdata}[2]{\expandafter\newcommand\csname data-#1\endcsname{#2}}

\newcommand{\cidS}{cfgB-kl-rb-rmb-rgsb-rgm-aa-ff-ca-zi-eq-nls-wn-fwn-lrd-ref70kb-rc-rr-res-pne-nnd-np-rb0.3-lag-log-bc-rd}
\newcommand{\cidLD}{b9-mc320-lr80-lim80-drop0.10}
\newcommand{\cidXXLD}{b9-mc448-lr65-lim65-drop0.10}

\definecolor{C0}{rgb}{0.121569, 0.466667, 0.705882}
\definecolor{C1}{rgb}{1.000000, 0.498039, 0.054902}
\definecolor{C2}{rgb}{0.172549, 0.627451, 0.172549}
\definecolor{C3}{rgb}{0.839216, 0.152941, 0.156863}
\definecolor{C4}{rgb}{0.580392, 0.403922, 0.741176}
\definecolor{C5}{rgb}{0.549020, 0.337255, 0.294118}
\definecolor{C6}{rgb}{0.890196, 0.466667, 0.760784}
\definecolor{C7}{rgb}{0.498039, 0.498039, 0.498039}
\definecolor{C8}{rgb}{0.737255, 0.741176, 0.133333}
\definecolor{C9}{rgb}{0.090196, 0.745098, 0.811765}
\definecolor{C10}{rgb}{1,1,1}
\definecolor{C11}{rgb}{0,0,0}

\input{data/data-edm2}
\defdata{img512-ADM-fid2}{23.24}
\defdata{img512-ADM-cfg2}{7.72}
\defdata{img512-ADM-mparams0}{559}
\defdata{img512-ADM-gflops0}{1983}
\defdata{img512-ADM-nfe}{250}
\defdata{img512-ADM-mimg0}{497}
\defdata{img512-ADM-U-fid2}{9.96}
\defdata{img512-ADM-U-cfg2}{3.85}
\defdata{img512-ADM-U-mparams0}{730}
\defdata{img512-ADM-U-gflops0}{2813}
\defdata{img512-ADM-U-nfe}{250}
\defdata{img512-ADM-U-mimg0}{361}

\defdata{img512-ADM-release}{2021.42}
\defdata{img512-ADM-U-release}{2021.42}

\defdata{img512-DiT-XL-fid2}{12.03}
\defdata{img512-DiT-XL-cfgnointerval}{3.04}
\defdata{img512-DiT-XL-cfg2}{2.40}
\defdata{img512-DiT-XL-mparams0}{675}
\defdata{img512-DiT-XL-gflops0}{525}
\defdata{img512-DiT-XL-nfe}{250}
\defdata{img512-DiT-XL-mimg0}{768}

\defdata{img512-RIN-fid2}{3.95}
\defdata{img512-RIN-cfg2}{3.95}  
\defdata{img512-RIN-mparams0}{320}
\defdata{img512-RIN-gflops0}{415}
\defdata{img512-RIN-nfe}{1000}
\defdata{img512-RIN-mimg0}{1024}
\defdata{img512-RIN-release}{2022.92}

\defdata{img512-U-ViT-L-fid2}{3.54}
\defdata{img512-U-ViT-L-cfg2}{3.02}
\defdata{img512-U-ViT-L-mparams0}{2455}
\defdata{img512-U-ViT-L-gflops0}{555}
\defdata{img512-U-ViT-L-nfe}{256}
\defdata{img512-U-ViT-L-mimg0}{1024}
\defdata{img512-U-ViT-L-release}{2023.08}

\defdata{img512-GI-S-cfg2}{1.68}
\defdata{img512-GI-S-gflops0}{102}
\defdata{img512-GI-S-mimg0}{2147}

\defdata{img512-GI-XXL-cfg2}{1.40}
\defdata{img512-GI-XXL-gflops0}{552}
\defdata{img512-GI-XXL-mimg0}{940}

\defdata{img512-VDMpp-fid2}{2.99}
\defdata{img512-VDMpp-cfg2}{2.65}
\defdata{img512-VDMpp-mparams0}{2455}
\defdata{img512-VDMpp-gflops0}{555}
\defdata{img512-VDMpp-nfe}{256}
\defdata{img512-VDMpp-mimg0}{1434}
\defdata{img512-VDMpp-release}{2023.25}

\defdata{img512-StyleGAN-XL-fid2}{--}
\defdata{img512-StyleGAN-XL-cfg2}{2.41}
\defdata{img512-StyleGAN-XL-mparams0}{168}
\defdata{img512-StyleGAN-XL-gflops0}{2067}
\defdata{img512-StyleGAN-XL-nfe}{1}
\defdata{img512-StyleGAN-XL-mimg0}{640}

\defdata{img512-VAE-gflops0}{1261}
\defdata{img512-VAE-gflops1}{1260.9}

\defdata{img512-Pagoda-cfg2}{1.80}
\defdata{img512-Pagoda-time}{0.046}

\defdata{img512-SD2-small-250-cfg2}{3.55}
\defdata{img512-SD2-small-400-cfg2}{2.75}
\defdata{img512-SD2-small-750-cfg2}{2.33}
\defdata{img512-SD2-small-1000-cfg2}{2.19}
\defdata{img512-SD2-small-cfg2}{2.19}
\defdata{img512-SD2-small-gflops0}{137}
\defdata{img512-SD2-small-mimg0}{1000 * 2048 / 1e3}
\defdata{img512-SD2-small-cfgnointerval}{2.77}

\defdata{img512-SD2-flop-100-cfg2}{4.92}
\defdata{img512-SD2-flop-250-cfg2}{2.06}
\defdata{img512-SD2-flop-500-cfg2}{1.60}
\defdata{img512-SD2-flop-800-cfg2}{1.48}

\defdata{img512-SD2-flop-cfgnointerval}{1.99}
\defdata{img512-SD2-flop-cfg2}{1.48}
\defdata{img512-SD2-flop-gflops0}{653}
\defdata{img512-SD2-flop-mimg0}{1638}
\defdata{img512-SD2-flop-release}{2024.58}

\defdata{img512-SD2-distill-cfg2}{1.5}
\defdata{img512-SD2-distill-time}{0.29}

\defdata{img512-SD2-flop-pretrain-50-cfg2}{1.69}
\defdata{img512-SD2-flop-pretrain-100-cfg2}{1.63}
\defdata{img512-SD2-flop-pretrain-200-cfg2}{1.55}
\defdata{img512-SD2-flop-pretrain-cfg2}{1.55}
\defdata{img512-SD2-flop-pretrain-gflops0}{653}

\defdata{img512-SD2-flop-pretrain-pretrainflops}{400 * 2048 / 1e3 * 137 * 3 / 1e6}

\usepackage{amsmath,amsfonts,bm}

\def\eqref#1{equation~\ref{#1}}

\def\1{\bm{1}}

\def\eps{{\epsilon}}

\def\rvx{{\mathbf{x}}}

\def\veps{{\bm{\eps}}}

\def\vx{{\bm{x}}}

\def\vz{{\bm{z}}}

\DeclareMathAlphabet{\mathsfit}{\encodingdefault}{\sfdefault}{m}{sl}
\SetMathAlphabet{\mathsfit}{bold}{\encodingdefault}{\sfdefault}{bx}{n}

\newcommand{\sigmoid}{\sigma}

\def\colorprevlatent{C0!40!white}
\def\colorprev{C0}

\def\colorours{C2}
\def\coloroursextra{C2!40!white}

\newcommand{\figQualityComputeScatter}{%
\begin{figure*}[t]%
\centering\footnotesize%
\begin{tikzpicture}%
\begin{axis}[
  width={.95\textwidth}, height={80mm}, grid={major},
  xmin={33}, xmax={3200}, x coord trafo/.code=\pgfmathparse{##1^0.2}, xtick={50, 100, 200, 500, 1000, 2000}, xticklabels={$50$, $100$, $200$, $500$, $1000$, $2000$},
  ymin={.9}, ymax={15}, y coord trafo/.code=\pgfmathparse{ln(##1)}, ytick={1, 2, 3, 5, 10, 20, 30}, yticklabels={$1$, $2$, $3$, $5$, $10$, $20$, \raisebox{-1.5ex}[0ex][0ex]{FID}},
  legend pos={north west}, legend cell align={left},
]
\gdef\did{img512}
\gdef\pdotSize{1.5pt}

\gdef\pmx##1{\addplot[black, opacity=0.08, thin, forget plot] coordinates {(##1,.9) (##1,30)};}
\gdef\pmy##1{\addplot[black, opacity=0.08, thin, forget plot] coordinates {(33,##1) (3200,##1)};}
\pmx{40}\pmx{60}\pmx{70}\pmx{80}\pmx{90}\pmx{300}\pmx{400}\pmx{600}\pmx{700}\pmx{800}\pmx{900}\pmx{3000}
\pmy{1.1}\pmy{1.2}\pmy{1.3}\pmy{1.4}\pmy{1.5}\pmy{1.6}\pmy{1.7}\pmy{1.8}\pmy{1.8}\pmy{1.9}\pmy{2.1}\pmy{2.2}\pmy{2.3}\pmy{2.4}\pmy{2.5}\pmy{2.6}\pmy{2.7}\pmy{2.8}\pmy{2.9}\pmy{4}\pmy{6}\pmy{7}\pmy{8}\pmy{9}

\gdef\pcNoGuid##1{(\rawdata{\did-##1-gflops0}, \rawdata{\did-##1-fid2})}
\gdef\pcPrevGuid##1{(\rawdata{\did-##1-gflops0}, \rawdata{\did-##1-cfg2})}
\gdef\pcOurGuid##1{(\rawdata{\did-##1-gflops0}, \rawdata{\did-##1-fid2})}
\gdef\pcOurNoInterval##1{(\rawdata{\did-##1-gflops0}, \rawdata{\did-##1-cfgnointerval})}

\gdef\pdot##1##2##3##4{\addplot[##1, mark=*, mark size=\pdotSize, forget plot, nodes near coords align={##2}, nodes near coords=\textlabel{##3}] coordinates {##4};}
\pdot{\colorprev}{east}{ADM}{\pcPrevGuid{ADM}}
\pdot{\colorprev}{east}{ADM-U}{\pcPrevGuid{ADM-U}}
\pdot{\colorprevlatent}{east}{DiT-XL/2 (interval)}{\pcPrevGuid{DiT-XL}}
\pdot{\colorprev}{west}{U-ViT, L}{\pcNoGuid{U-ViT-L}}
\pdot{\colorprev}{west}{VDM++}{\pcPrevGuid{VDMpp}}
\pdot{\colorprevlatent}{north}{EDM2-S (interval)}{\pcPrevGuid{GI-S}}
\pdot{\colorprevlatent}{north}{EDM2-XXL (interval)}{\pcPrevGuid{GI-XXL}}
\pdot{\colorours}{north}{Small}{\pcPrevGuid{SD2-small}}
\pdot{\coloroursextra}{north}{Small}{\pcOurNoInterval{SD2-small}}
\pdot{\colorours}{south}{Flop Heavy}{\pcPrevGuid{SD2-flop}}
\pdot{\coloroursextra}{south}{Flop Heavy}{\pcOurNoInterval{SD2-flop}}
\addlegendimage{\colorprevlatent, only marks, mark size=\pdotSize}\addlegendentry{Previous (latent)}
\addlegendimage{\colorprev, only marks, mark size=\pdotSize}\addlegendentry{Previous (pixel)}
\addlegendimage{\coloroursextra, mark=*, only marks, mark size=\pdotSize}\addlegendentry{Ours (no interval)}
\addlegendimage{\colorours, mark=*, only marks, mark size=\pdotSize}\addlegendentry{Ours}

\end{axis}
\end{tikzpicture}%
\caption{\label{figQualityComputeScatter}%
Model forward pass complexity (gigaflops) on ImageNet512, figure adapted from \citep{karras2023edm2}. \textit{Note: both axis are in log-scale.} See also Table~\ref{tab:sampling_comparison} for more details.
}%
\end{figure*}
}%

\newcommand{\figQualityTrainingScatter}{%
\begin{figure*}[t]%
\centering\footnotesize%
\begin{tikzpicture}%
\def\CB{C2!70!black}
\def\CC{C2!70!black}
\begin{axis}[
  width={.99\linewidth}, height={60mm}, grid={major},
  xmin={0}, xmax={3.5}, xmode={linear}, xtick={0.0, 0.5, 1.0, 1.5, 2.0, 2.5, 3.0, 3.5}, xticklabels={$0.0$, $0.5$, $1.0$, $1.5$, $2.0$, $2.5$, $3.0$, $3.5$},
  ymin={.9}, ymax={11}, y coord trafo/.code=\pgfmathparse{ln(##1)}, ytick={1, 2, 3, 5, 10}, yticklabels={$1$, $2$, $3$, $5$, $10$, \raisebox{-1.5ex}[0ex][0ex]{FID}},
  xlabel={Training cost (zettaflops)}, x label style={at={(axis description cs:0.5,0.02)}, anchor=north},
  legend pos={north west}, legend cell align={left},
]
\gdef\did{img512}
\gdef\pdotSize{1.5pt}

\gdef\pmy##1{\addplot[black, opacity=0.08, thin, forget plot] coordinates {(0,##1) (3.5,##1)};}
\pmy{1.1}\pmy{1.2}\pmy{1.3}\pmy{1.4}\pmy{1.5}\pmy{1.6}\pmy{1.7}\pmy{1.8}\pmy{1.9}\pmy{2.1}\pmy{2.2}\pmy{2.3}\pmy{2.4}\pmy{2.5}\pmy{2.6}\pmy{2.7}\pmy{2.8}\pmy{2.9}\pmy{4}\pmy{6}\pmy{7}\pmy{8}\pmy{9}

\gdef\pcFinal##1{(\rawdata{\did-##1-mimg0} * \rawdata{\did-##1-gflops0} * 3 / 1e6, \rawdata{\did-##1-cfg2})}
\gdef\pcFinalNoG##1{(\rawdata{\did-##1-mimg0} * \rawdata{\did-##1-gflops0} * 3 / 1e6, \rawdata{\did-##1-fid2})}
\gdef\pcEDM##1##2{(##2 * \rawdata{\did-##1-gflops0} * 3 / 1e6, \rawdata{\did-##1-phema-##2-fid2})}
\gdef\pcSD##1##2{(##2 * 2048 / 1e3 * \rawdata{\did-##1-gflops0} * 3 / 1e6, \rawdata{\did-##1-##2-cfg2})}

\gdef\pcFinalNoInterval##1{(\rawdata{\did-##1-mimg0} * \rawdata{\did-##1-gflops0} * 3 / 1e6, \rawdata{\did-##1-cfgnointerval})}

\gdef\pcSDpretrain##1##2{(##2 * 2048 / 1e3 * \rawdata{\did-##1-gflops0} * 3 / 1e6 + \rawdata{\did-##1-pretrainflops}, \rawdata{\did-##1-##2-cfg2})}

\addplot[\CB, forget plot] coordinates {\pcSD{SD2-flop}{100}\pcSD{SD2-flop}{250}\pcSD{SD2-flop}{500}\pcSD{SD2-flop}{800}};
\addplot[\CB, forget plot] coordinates {\pcSD{SD2-small}{250}\pcSD{SD2-small}{400}\pcSD{SD2-small}{750}\pcSD{SD2-small}{1000}};
\addplot[\CB, forget plot] coordinates {\pcSDpretrain{SD2-flop-pretrain}{50}\pcSDpretrain{SD2-flop-pretrain}{100}\pcSDpretrain{SD2-flop-pretrain}{200}};

\gdef\pdot##1##2##3##4{\addplot[##1, mark=*, mark size=\pdotSize, forget plot, nodes near coords align={##2}, nodes near coords=\textlabel{##3}] coordinates {##4};}
\gdef\pdott##1##2##3##4{\addplot[opacity=0.8, forget plot, nodes near coords align={##2}, nodes near coords=\textlabel{##3}] coordinates {##4};}

\pdot{\colorprev}{east}{ADM}{\pcFinal{ADM}}
\pdot{\colorprev}{east}{ADM-U}{\pcFinal{ADM-U}}
\pdot{\colorprevlatent}{south}{DiT-XL/2 (interval)}{\pcFinal{DiT-XL}}
\pdot{\colorprev}{west}{RIN}{\pcFinalNoG{RIN}}
\pdot{\colorprev}{west}{SD, U-ViT-L}{\pcFinal{U-ViT-L}}
\pdot{\colorprev}{west}{VDM++}{\pcFinal{VDMpp}}
\pdot{\colorprevlatent}{west}{EDM2-S (interval)}{\pcFinal{GI-S}}
\pdot{\colorprevlatent}{west}{EDM2-XXL (interval)}{\pcFinal{GI-XXL}}
\pdot{\colorprevlatent}{south}{EDM2-S}{\pcFinalNoG{\cidS}}
\pdot{\colorprevlatent}{south}{EDM2-L}{\pcFinalNoG{\cidLD}}
\pdot{\colorprevlatent}{west}{EDM2-XXL}{\pcFinalNoG{\cidXXLD}}

\pdot{\coloroursextra}{south west}{(no interval)}{\pcFinalNoInterval{SD2-small}}
\pdot{\CB}{north}{Small}{\pcFinal{SD2-small}}
\pdot{\coloroursextra}{south}{(no interval)}{\pcFinalNoInterval{SD2-flop}}
\pdot{\CB}{south}{Flop Heavy}{\pcFinal{SD2-flop}}
\pdot{\CB}{north}{Flop Heavy (finetune)}{\pcSDpretrain{SD2-flop-pretrain}{200}}

\addlegendimage{\colorprevlatent, only marks, mark size=\pdotSize}\addlegendentry{Previous (latent)}
\addlegendimage{\colorprev, only marks, mark size=\pdotSize}\addlegendentry{Previous (pixel)}
\addlegendimage{\CB, only marks, mark size=\pdotSize}\addlegendentry{Simpler diffusion (ours)}

\end{axis}
\end{tikzpicture}%
\vspace{-.2cm}
\caption{\label{figQualityTrainingScatter}%
FID versus training cost on ImageNet512.
Note that the EDM-2 training curves are without guidance, but the best performance with guidance intervals is reported with a separate marker, the figure source adapted has been adapted from EDM2 \citep{karras2023edm2}. Following their assumptions, one training iteration is three times as expensive as evaluating the model. Our Flop Heavy (finetune) is trained by first training a small model on ImageNet256 for 400k iterations and then finetuning the same model on ImageNet512.
}%
\end{figure*}
}%

\newcommand{\figEndDiffusionOverYears}{%
\begin{figure}
\centering\footnotesize%
\begin{tikzpicture}%
\begin{axis}[
  width={.99\linewidth}, height={40mm}, grid={major},
  xmin={2020.9}, xmax={2024.9}, xtick={2021, 2022, 2023, 2024}, xticklabels={$2021$, $2022$, $2023$, $2024$},
  ymin={.9}, ymax={9}, y coord trafo/.code=\pgfmathparse{ln(##1)}, ytick={1, 2, 3, 5, 9}, yticklabels={$1$, $2$, $3$, $5$, \raisebox{-2ex}[0ex][0ex]{FID}},
  legend pos={north west}, legend cell align={left},
]
\gdef\did{img512}
\gdef\pdotSize{1.5pt}
\gdef\pcTime##1{(\rawdata{\did-##1-release}, \rawdata{\did-##1-cfg2})}
\gdef\pdot##1##2##3##4{\addplot[##1, mark=*, mark size=\pdotSize, forget plot, nodes near coords align={##2}, nodes near coords=\textlabel{##3}] coordinates {##4};}
\pdot{\colorprev}{west}{ADM}{\pcTime{ADM}}
\pdot{\colorprev}{east}{RIN}{\pcTime{RIN}}
\pdot{\colorprev}{west}{SD, U-ViT L}{\pcTime{U-ViT-L}}
\pdot{\colorprev}{north}{VDM++}{\pcTime{VDMpp}}
\pdot{\colorours}{north east}{SiD2 (ours)}{\pcTime{SD2-flop}}
\end{axis}
\end{tikzpicture}%
\caption{\label{fig:end-to-end-years}%
End-to-end pixel diffusion performance on ImageNet512.
}%
\vspace{-0.5cm}%
\end{figure}
}%

\newcommand{\figLosses}{
    \begin{figure*}
    \begin{minipage}[t]{.49\textwidth}
        \centering
        \includegraphics[width=\textwidth]{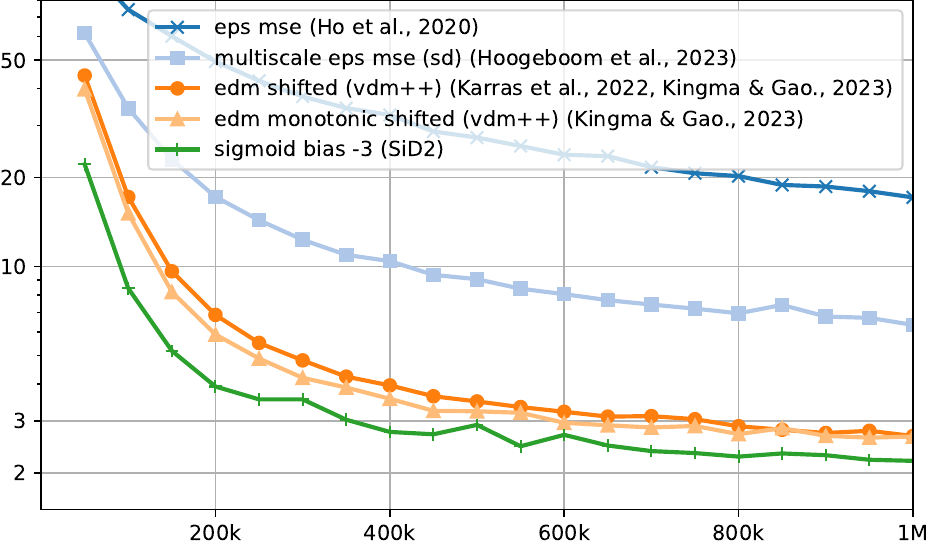}
        \caption{Effect of weighting functions in the loss on the FID score of ImageNet512 over the training iterations. The proposed sigmoid weighting with $b\!=\!-3$ shows best performance consistent across all training iterations.
        }
        \label{fig:loss_weighting}
    \end{minipage}%
    \hfill
    \begin{minipage}[t]{.49\textwidth}
        \centering
        \includegraphics[width=\textwidth]{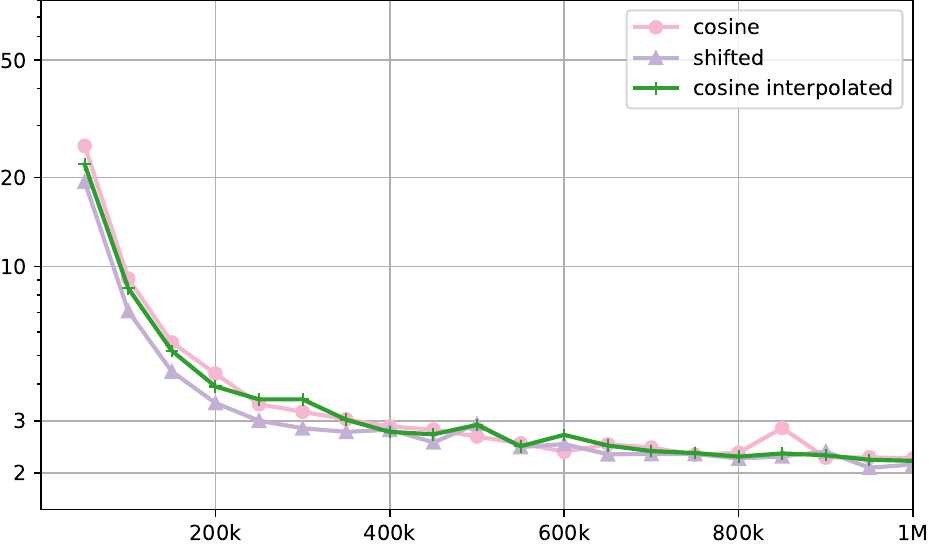}
        \caption{Effect of \textit{training} noise schedules on the FID score of ImageNet512 over the training iterations. Fortunately, with our proposed sigmoid weighting ($b\!=\!-3$), the exact choice of the noise schedules has little effect on the \textit{training} process.
        }
        \label{fig:sample_efficiency}
    \end{minipage}
    \vspace{-.4cm}
    \end{figure*}
}

\newcommand{\tablemscoco}{%
\begin{table}
\centering
\caption{Zero-shot text-to-image performance evaluated on MSCOCO in FID$_{\text{30k}}$.
Our distilled model, evaluated on the more difficult 512 setting, outperforming other works.
}
\label{tab:literature_tti}
\scalebox{.87}{
\begin{tabular}{@{}llrr@{}}
\toprule
Method & Size & NFE & FID$_{\text{30k}}$ \\ \midrule
DeepFloyd \citepsmall{deepfloyd2024github} & 256 & & 6.7\\
eDiff-I \citepsmall{balaji2022ediffi} & 256 & & 6.9 \\ 
Parti \citepsmall{yu2022scalingautoregressiveparti} & 256 & & 7.2 \\	
Imagen \citepsmall{saharia2022imagen} & 256 & & 7.3 \\
simple diffusion \citepsmall{hoogeboom2023simple} & 256 &  & 8.3 \\ \midrule
simple diffusion \citepsmall{hoogeboom2023simple} & 512 & 256 & 9.6 \\
SDv1.5 (eval DMD2) \citepsmall{rombach2022highresolution} & 512 & 200 & 7.2 \\
DMD2 \citepsmall{yin2024dmd2} & 512 & 4 & 8.3 \\
\midrule
SiD2 (ours, bias=-3) & 512 & 256 & 8.1 \\
SiD2 16-step distilled (ours) & 512 & 16 & \textbf{6.7} \\
\bottomrule
\end{tabular}} \vspace{-.2cm}
\end{table}}

\newcommand{\tablekinetics}{%
\begin{table}
\centering
\caption{Conditional Video generation performance on Kinetics600 in FVD (resized to 64). For video generation the loss specification is important to obtain good sample quality (see \textit{ablations}).
}
\label{tab:literature_kinetics}
\scalebox{.84}{
\begin{tabular}{@{}lr@{}}
\toprule
Method & FVD$_{\text{50k}}$ \\ \midrule
\textit{pixel diffusion} \\
Video Diffusion Models \citep{ho2022videodiffusion} & 16.6 \\ 
RIN  \citep{jabri2022scalable} & 10.8 \\
Rolling Diffusion \citep{ruhe2024rolling} & 5.2 \\ \midrule
\textit{latent diffusion} \\
MAGVIT \citep{yu2022magvit} & 9.9 \\
MAGVITv2 \citep{yu2024magvitv2} & 4.3 \\
W.A.L.T \citep{gupta2024walt} & 3.3 \\
\midrule
SiD2 (ours, sigmoid bias = 0) & \textbf{2.3} \\ 
\textit{ablations} \\
\hspace{.6cm}eps-mse & 5.7 \\ 
\hspace{.6cm}sigmoid bias = +2 & 3.2 \\
\hspace{.6cm}sigmoid bias = -2 & 2.5 \\
\bottomrule
\end{tabular}}
\vspace{-.2cm}
\end{table}}

\newcommand{\figScaling}{
    \begin{figure}
    \centering
    \includegraphics[width=.47\textwidth]{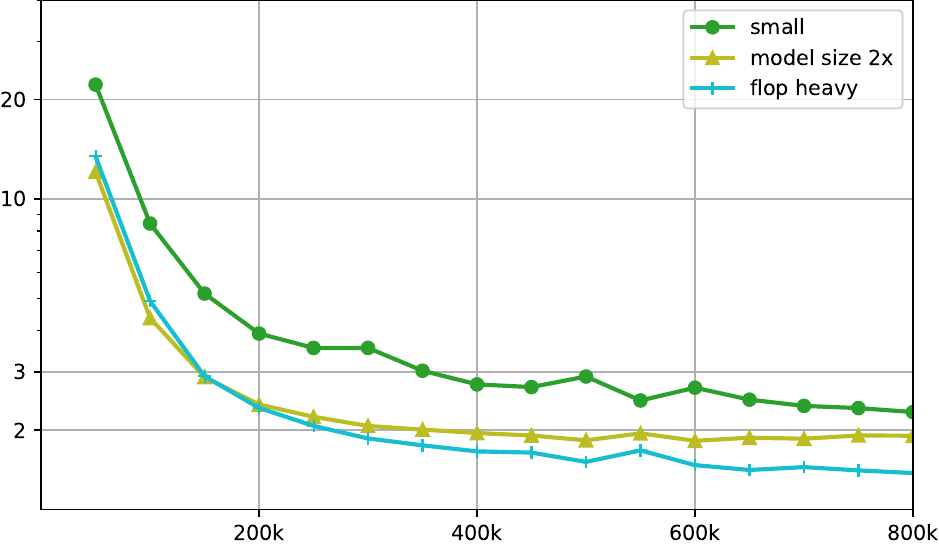}\vspace{.2cm}
    \caption{The effect of scaling either channels or tokens, compared to the small model variant. FLOPS between model size 2x and flop heavy are similar.
    }
    \label{fig:scaling}
    \end{figure}
    \begin{figure}
    \centering
    \includegraphics[width=.47\textwidth]{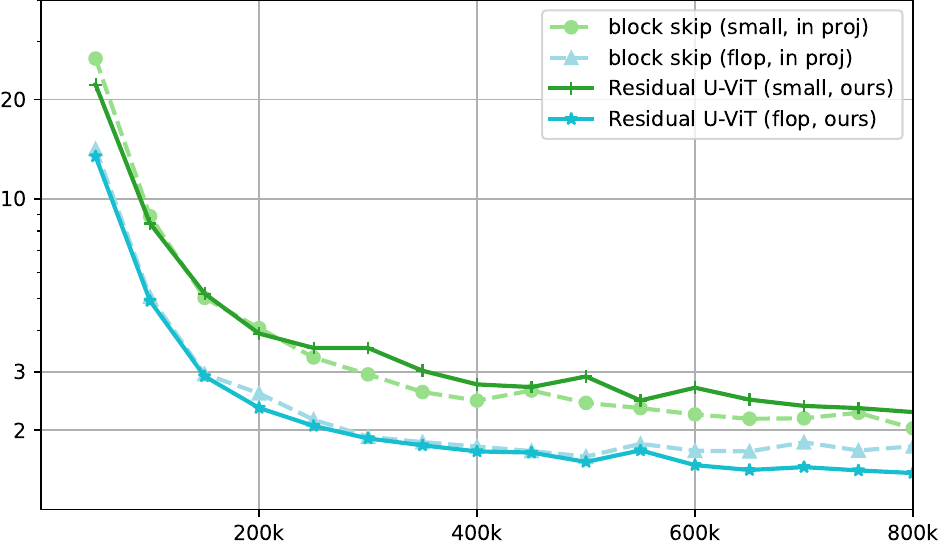}
    \caption{The effect of Residual-UViTs (removing block skip connections). Although smaller models can be effected, larger models are unaffected by the simplification.
    }
    \label{fig:skip}
    \end{figure}
}

\newcommand{\figGrid}{
    \begin{table}[t]
    \caption{
    Sigmoid weighting performance (\textit{left}) with the \textit{flop heavy} architecture. Larger resolutions require more negative bias.
    Power loss performance (\textit{right}) on ImageNet512 with the \textit{small model}. Increasing downsampling has a similar effect as decreasing bias, hence we prefer the more simple sigmoid weighting.
    }    
    \begin{subtable}[t]{.49\columnwidth}
        \caption{Sigmoid weighting}
        \label{tab:sigmoid_resolutions}
        \resizebox{!}{17mm}{
            \begin{tabular}{ccccc}
                \toprule
                \multicolumn{1}{c}{} & \multicolumn{4}{c}{\textbf{resolution}} \\
                \cmidrule(rl){2-5}
                \textbf{bias} & {128} & {256} & {512} & {1024} \\
                \midrule
                 -5 & - & - & - & 1.82  \\
                 -4 & - & - & 1.63 & \textbf{1.75}  \\
                 -3 & - & 1.47 & \textbf{1.48} & 2.32\\
                 -2 & 1.49 & 1.42 & 1.55 & 2.17 \\
                 -1 & 1.35 & \textbf{1.38} & 1.75 & -   \\
                 0 & 1.30 & 1.46 & 1.94 & -  \\
                 1 & \textbf{1.26} & 1.56 & - & -  \\
                \bottomrule
            \end{tabular}
        }
    \end{subtable}%
    \hfill
    \begin{subtable}[t]{.49\columnwidth}
        \caption{Power loss weighting}
        \label{tab:power_loss}
        \resizebox{!}{17mm}{
            \begin{tabular}{ccccc}
            \toprule
            \multicolumn{1}{c}{} & \multicolumn{4}{c}{\textbf{downsample levels}} \\
            \cmidrule(rl){2-5}
            \textbf{bias} & {0} & {1} & {2} & {3} \\
            \midrule
            -4 &  \textbf{2.19} & 2.58 & 3.62 & 5.11 \\
            -3 &  \textbf{2.19} & 2.32 & 2.85 & 4.02 \\
            -2 & 2.29 &  \textbf{2.15} & 2.34 & 3.13 \\
            -1 & 2.64 & 2.32 & 2.25 & 2.54 \\
             0 & 3.22 & 2.74 &  \textbf{2.07} & 2.22 \\
             1 &  - & 2.69 & 2.29 &  \textbf{2.16} \\
             2 &  - & 3.26 & 2.53 & 2.19 \\
            \bottomrule
            \end{tabular}
        }
    \end{subtable}
    \vspace{-.2cm}
\end{table}
}

\newcommand{\figNoGuidanceLoss}{
\begin{figure}
    \centering
    \includegraphics[width=.44\textwidth]{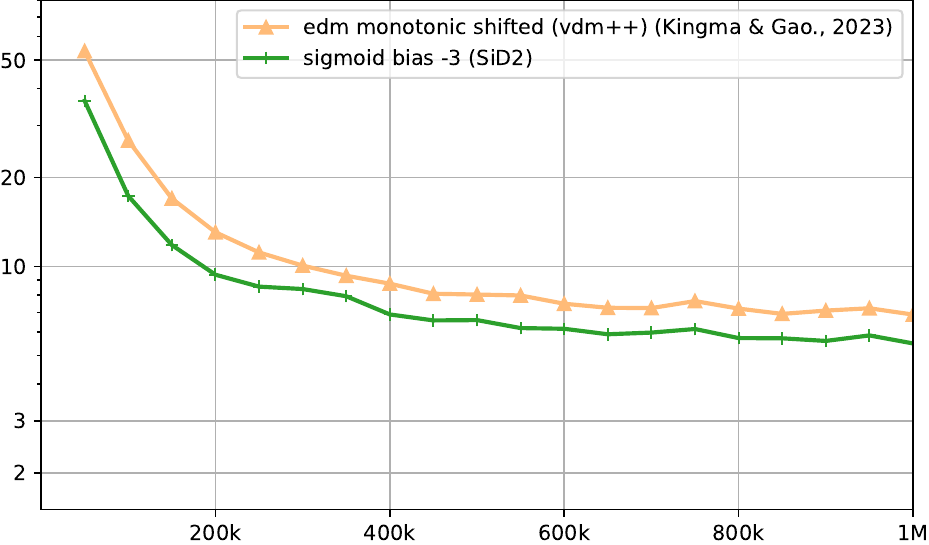}
    \caption{The sigmoid $b$ $=$ $-3$ loss versus the EDM monotonic loss, without any guidance with the small architecture variant.}
    \label{fig:noguidance}
\end{figure}}

\newcommand{\figExamples}{
\begin{figure*}
    \centering
    \includegraphics[width=.24\textwidth]{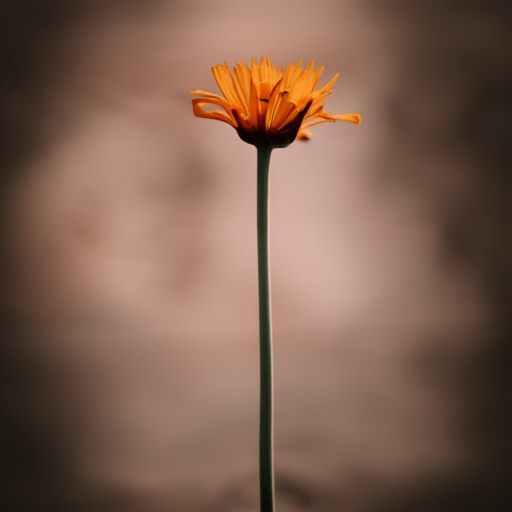} \hfill
    \includegraphics[width=.24\textwidth]{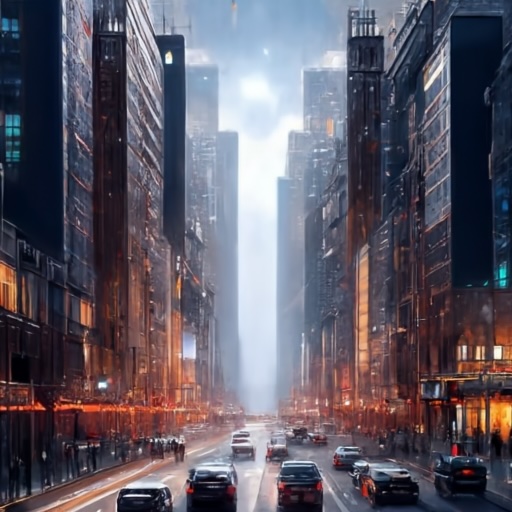} \hfill
    \includegraphics[width=.24\textwidth]{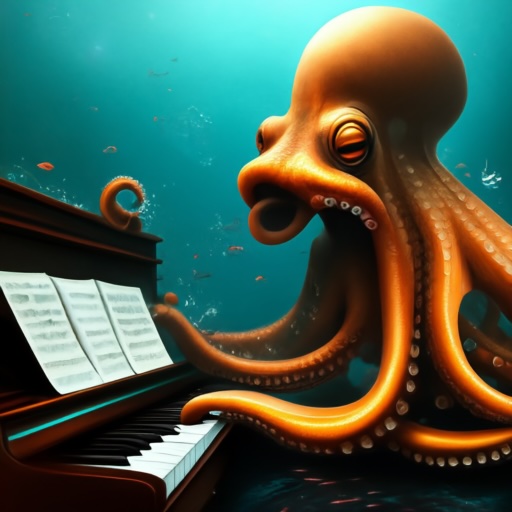} \hfill
    \includegraphics[width=.24\textwidth]{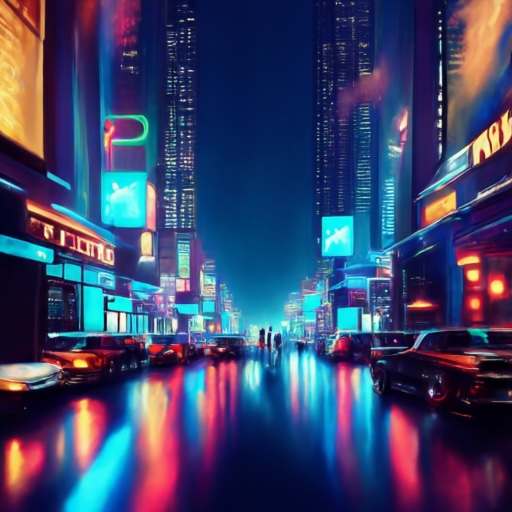}
     \caption{Some text-to-image samples of $512 \times 512$ that have been generated by SiD2.}
    \label{fig:examples}
\end{figure*}}

\newcommand{\figEncoderDecoderAppendix}{
    \begin{figure*}
        \centering
        \begin{subfigure}[b]{\columnwidth}
            \centering
            \includegraphics[width=.95\columnwidth]{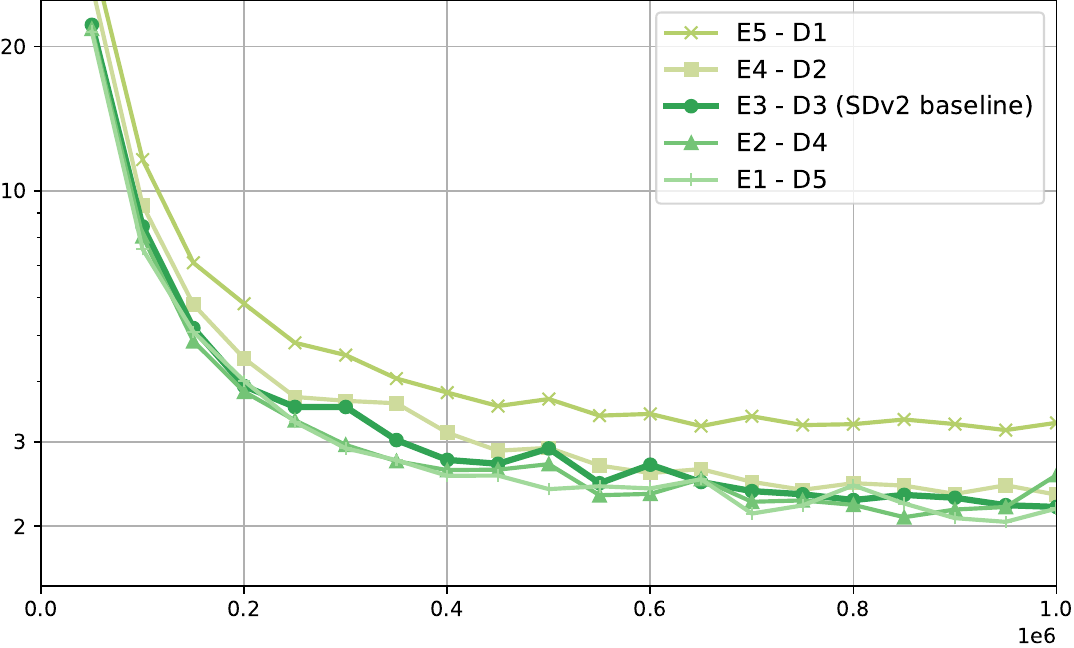}
            \caption{fixed depth}
            \label{fig:exp_encoder_decoder_fixed_params}
        \end{subfigure}
        \begin{subfigure}[b]{\columnwidth}
            \centering
            \includegraphics[width=.95\columnwidth]{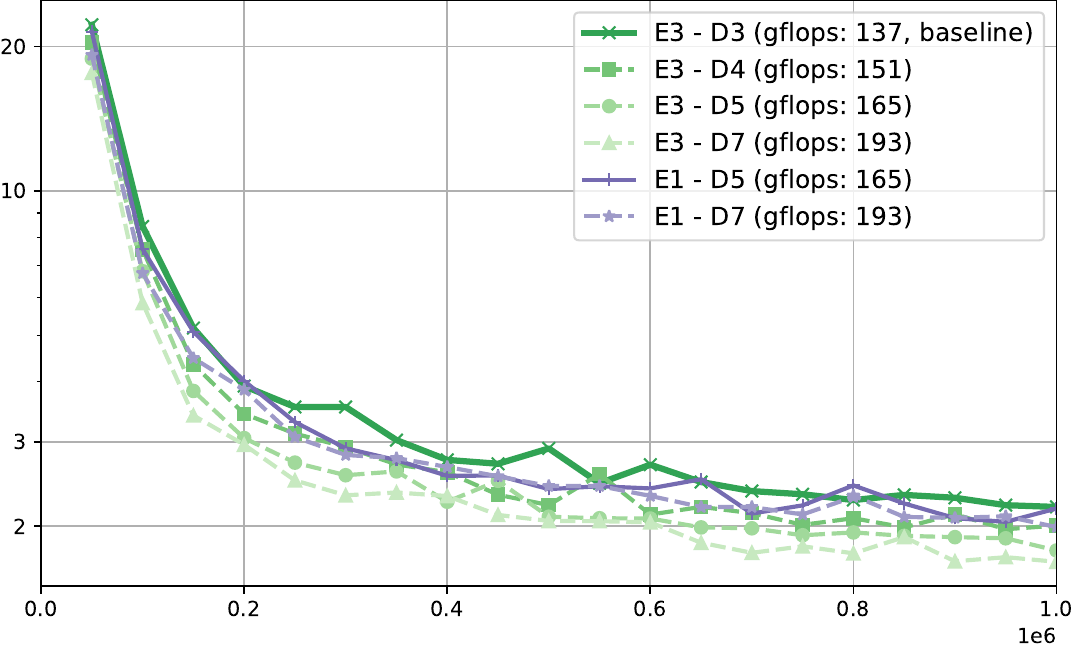}
            \caption{increasing depth}
            \label{fig:exp_encoder_decoder_increase_params}
        \end{subfigure}
        \caption{Encoder-Decoder Experiments}
        \label{fig:exp_asymmetric_uvit}
    \end{figure*}
}

\newcommand{\figTimeShiftedSigmoidAppendix}{
    \begin{figure}
        \centering
        \begin{subfigure}[b]{0.48\textwidth}
            \centering
            \includegraphics[width=.95\textwidth]{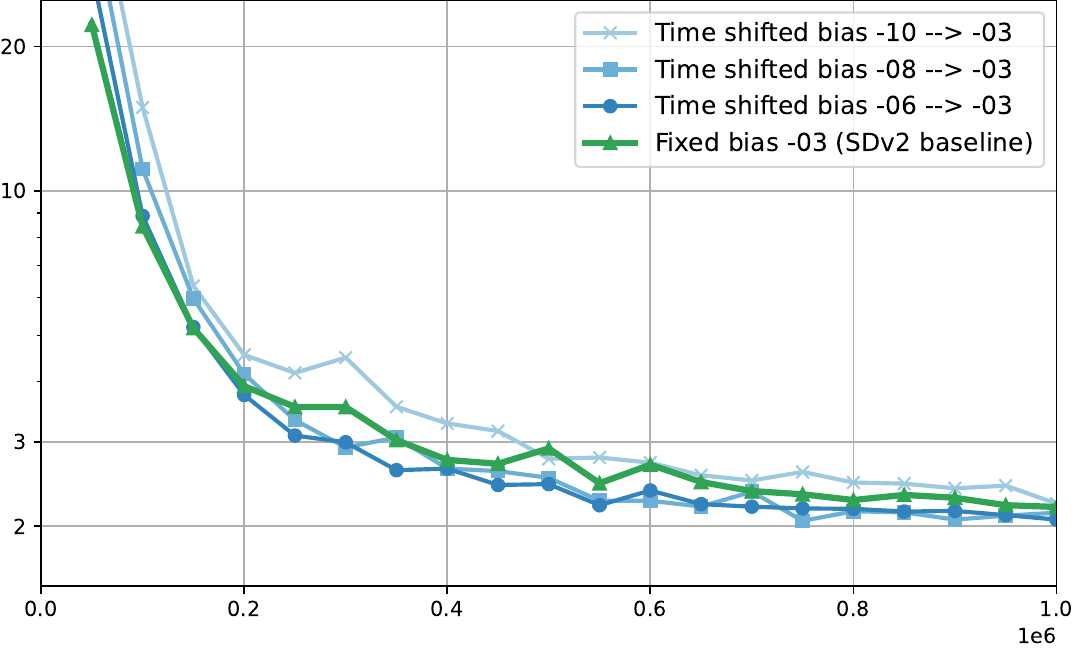}
            \caption{100k}
            \label{fig:exp_timeshift_100k}
        \end{subfigure}
        \hfill
        \begin{subfigure}[b]{0.48\textwidth}
            \centering
            \includegraphics[width=.95\textwidth]{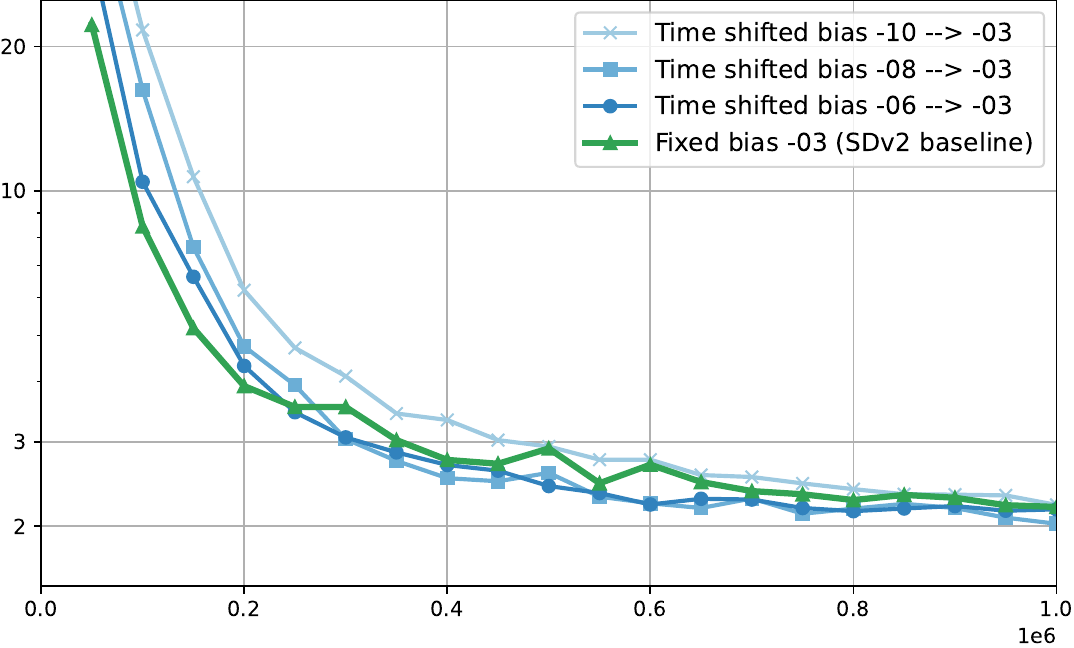}
            \caption{200k}
            \label{fig:exp_timeshift_200k}
        \end{subfigure}
        \caption{Timeshift results. Setting the bias more aggressively and then annealing to the best known setting improves performance somewhat, although the differences are relatively small.}
        \label{fig:exp_timeshift_all}
    \end{figure}
}

\newcommand{\figScalingAndEncoderDecoder}{
    \begin{figure*}
    \begin{minipage}[t]{.325\textwidth}
        \centering
        \includegraphics[width=\textwidth]{images/scaling.pdf}
        \caption{The effect of scaling either channels or tokens, compared to the small model variant. FLOPS between model size 2x and flop heavy are similar.}    
        \label{fig:scaling}
    \end{minipage}%
    \hfill
    \begin{minipage}[t]{.325\textwidth}
        \centering
        \includegraphics[width=\textwidth]{images/skip.pdf}
        \caption{The effect of Residual-UViTs (removing block skip connections). Although smaller models can be effected, larger models are unaffected by the simplification.}
        \label{fig:skip}
    \end{minipage}%
    \hfill
    \begin{minipage}[t]{.325\textwidth}
        \centering
        \includegraphics[width=\textwidth]{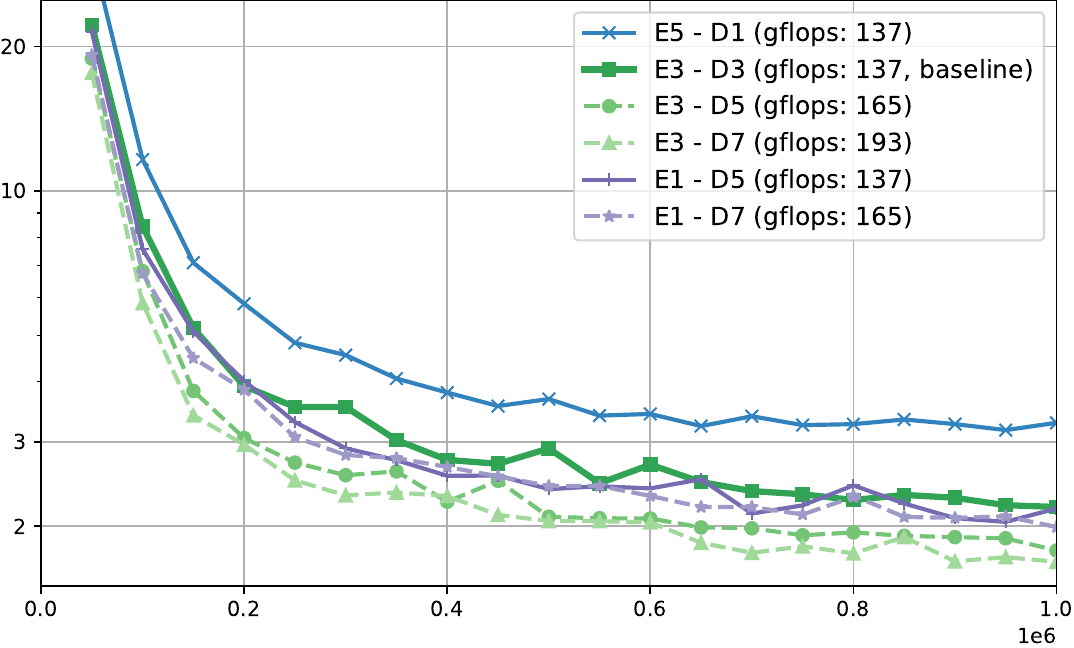}
        \caption{Changing the encoder decoder architecture. Decoder blocks are more important than encoder blocks, and symmetric architectures have decent performance.}
        \label{fig:exp_encoder_decoder}
    \end{minipage}
    \end{figure*}
}

\newcommand{\figLowBitLoss}{
    \begin{figure}
        \centering
        \includegraphics[width=.45\textwidth]{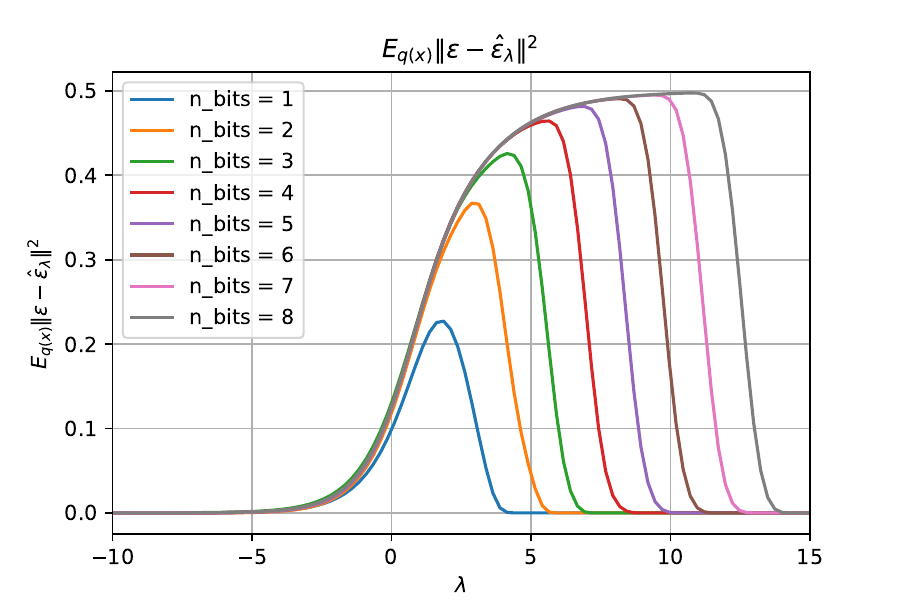}
        \caption{Expected loss over log signal-to-noise ratio $\lambda$ for multiple data distributions of different bit precision.}
        \label{fig:deps}
    \end{figure}
}

\newcommand{\figLowBitWeighted}{
    \begin{figure}
        \centering
        \includegraphics[width=.45\textwidth]{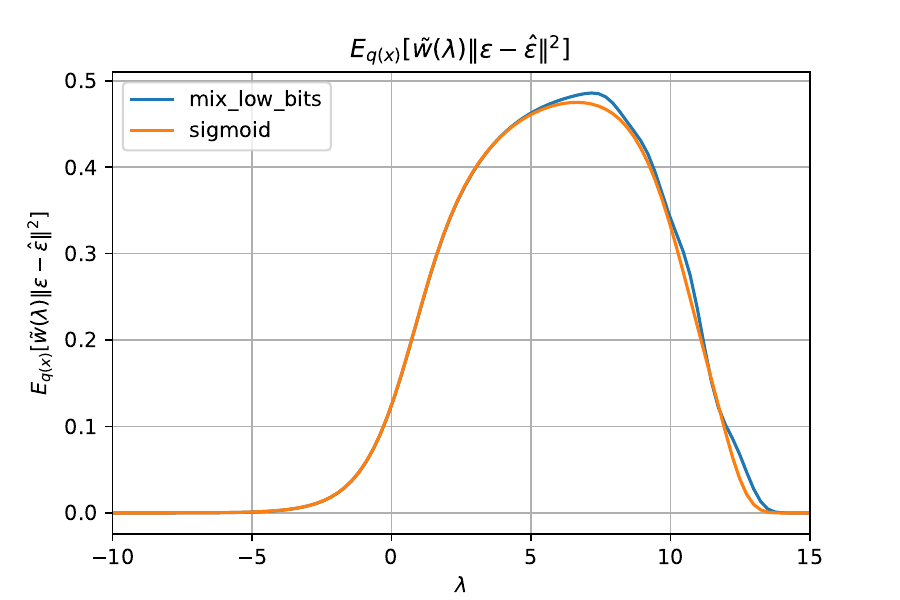}
        \caption{Weighted loss with a sigmoid weighting, compared with a low-bit training of a mixture of precisions.}
        \label{fig:mix_loss}
    \end{figure}
}

\renewcommand{\figScaling}{\figScalingAndEncoderDecoder}

\title{Simpler Diffusion (SiD2): \Large{1.5 FID on ImageNet512 with pixel-space diffusion}}

\author{%
  Emiel Hoogeboom \\
  Google Deepmind \\
  \texttt{emielh@google.com} \\
   \and
   Thomas Mensink \\
   Google Deepmind \\
     \texttt{mensink@google.com} \\
   \and
   Jonathan Heek \\
   Google Deepmind \\
   \texttt{jheek@google.com} \\
   \and
   Kay Lamerigts \\
   Google Research \\
   \texttt{kaylamerigts@google.com} \\
   \and
   Ruiqi Gao \\
   Google Deepmind \\
   \texttt{ruiqig@google.com}
   \and
   Tim Salimans \\
   Google Deepmind \\
   \texttt{salimans@google.com}
}

\begin{document}

\maketitle

\begin{abstract}
Latent diffusion models have become the popular choice for scaling up diffusion models for high resolution image synthesis. Compared to pixel-space models that are trained end-to-end, latent models are perceived to be more efficient and to produce higher image quality at high resolution. Here we challenge these notions, and show that pixel-space models can 
be very competitive to latent 
models both in quality and efficiency, achieving 1.5 FID on ImageNet512 and new SOTA results on ImageNet128, ImageNet256 and Kinetics600.

We present a simple recipe for scaling end-to-end pixel-space diffusion models to high resolutions. 1: Use the sigmoid loss-weighting \citep{kingma2023understandingdiffusion_vdmplus} with our prescribed hyper-parameters. 2: Use our simplified memory-efficient architecture with fewer skip-connections. 3: Scale the model to favor processing the image at a high resolution with fewer parameters, rather than using more parameters at a lower resolution.
Combining these with guidance intervals, we obtain a family of pixel-space diffusion models we call\\\emph{Simpler Diffusion} (SiD2).
\end{abstract}

\section{Introduction}
Diffusion models have been tremendously successful in domains such as image, video and audio generation. For high resolution data, \emph{latent diffusion models} \citep{rombach2022highresolution} have been particularly successful: Through encoding the data in a lower dimensional space, these models sidestep the difficulties faced by more traditional pixel-space diffusion models in scaling up to high resolution. This makes latent diffusion models also scale better in training compute than other approaches, as shown by \cite{karras2023edm2}. 

\figEndDiffusionOverYears
\figExamples
However, latent models have their own challenges: It can be difficult to train the autoencoders that produce the latents, which in practice requires careful tuning of regularization hyperparameters \citep{rombach2022highresolution}. Even when trained well, the decoder of a latent diffusion model can produce artifacts or otherwise imperfect reconstructions, especially when it is applied to latents that are out of distribution compared to the training set. A single-stage pixel-space diffusion model does not have these issues, and so we ask the question: Can pixel-space diffusion models scale as well as latent models with the right architecture and hyperparameters?

Over the years, end-to-end pixel-space diffusion has improved considerably (Figure~\ref{fig:end-to-end-years}).
The main problem was that the standard schedule and loss-weighting \citep{ho2020denoising} work well for small resolution images, like $32^2$ or $64^2$, but not at higher resolution \citep{chen2023importancenoise,hoogeboom2023simple}. It was further shown \citep{kingma2023understandingdiffusion_vdmplus} that it can be helpful to disentangle loss weighting and time sampling for improved control over both, leading to better performance at higher resolution. However, in previous works pixels-space models still did not match the best latent models on standard benchmarks like ImageNet512.

The main contribution of this paper is to improve pixel-space diffusion on high resolution images while keeping as close to the original formulation of diffusion models as possible. Specifically, we analyze and tune the sigmoid loss weighting from \citep{kingma2023understandingdiffusion_vdmplus}, and show why it should be preferred over their favored EDM monotonic. We further propose architectural improvements that lead to even better performance, namely flop heavy scaling and removing block skip-connections (Residual U-ViTs). Most notably, we improve the state-of-the-art of pixel-space diffusion from 2.65 to 1.48 on ImageNet512 (see also Figure~\ref{fig:end-to-end-years}).
Further, we also achieve overall state-of-the-art on ImageNet128, ImageNet256 and Kinetics600 without heavy tuning.

\subsection{Background}

A diffusion process gradually destroys data over a conceptual time $t$, typically with Gaussian noise as follows:
\begin{equation}
    \vz_t = \alpha_t \vx + \sigma_t \veps, \text{ where } \veps \sim \mathcal{N}(0, \mathbf{I}),
\end{equation}
which defines a distribution $q(\vz_t | \vx)$ such that $\vz_0$ is close to clean data and $\vz_1$ is close to Gaussian noise. In this paper we assume a variance-preserving~\citep{song2021scorebasedsde} schedule, which is defined as $\alpha_t = \sqrt{\sigma(\lambda_t)}$ and $\sigma_t = \sqrt{\sigma(-\lambda_t)}$, with $\lambda_t = \log \big{(} \alpha_t^2 / \sigma_t^2 \big{)}$ being the log signal-to-noise ratio, which monotonically decreases with time as $t \rightarrow 1$, and $\sigma(\cdot)$ being the sigmoid function. 
It can be shown that other types of schedules such as the variance-exploding schedule or the flow matching schedule~\citep{lipman2023flowmatching} can always be converted to a variance-preserving schedule with a linear scaling of $\vz_t$ \citep{kingma2023understandingdiffusion_vdmplus}. The key defining property of a noise schedule is the log signal-to-noise ratio $\lambda_t$.
A neural network is used to predict $\hat{\vx} = \hat{\vx}_\theta(\vz_t, t)$ that effectively estimates $\mathbb{E}[\vx | \vz_t]$, which can be done by minimizing a weighted mean squared error loss:
\begin{equation}
    L(\vx) = \mathbb{E}_{t \sim \mathcal{U}(0, 1)} \textcolor{gray}{-\frac{\mathrm{d}\lambda_t}{ \mathrm{d}t}} w(\lambda_t)  \| \vx - \hat{\vx} \|^2,
    \label{eq:loss}
\end{equation}
where $w(\lambda_t)$ is the weighting function. $\textcolor{gray}{-\mathrm{d}\lambda_t / \mathrm{d}t}$ cancels out the effect of the noise schedule $\lambda_t$ on the loss. In other words, $L(\rvx)$ can be rewritten as $\int - w(\lambda) \| \vx - \hat{\vx} \|^2 d\lambda$ that is invariant to the choice of $\lambda_t$. However, $\lambda_t$ still impacts the \textit{sampling efficiency} of the Monte Carlo estimator of the expected loss, and thus affects the training dynamics of diffusion models. This decouples \textit{noise schedule} $\lambda_t$ and \textit{weighting function} $w(\lambda)$ as a function of $\lambda$~\citep{kingma2023understandingdiffusion_vdmplus}.

The above squared error loss can be derived from the variational interpretation of diffusion models~\citep{kingma2021vdm}. Given two time steps $s < t$, we define the denoising distribution $p(\vz_s | \vz_t)$ in the same formulation as the true posterior $q(\vz_s | \vz_t, \vx)$, but replace the ground truth $\vx$ with the learned approximation $\hat{\vx} \approx \mathbb{E}[\vx | \vz_t]$:
\begin{equation} \small
    p(\vz_s | \vz_t) = q(\vz_s | \vz_t, \vx=\hat{\vx}) = \mathcal{N}(\vz_s | \mu_{t \to s}, \sigma_{t \to s}),
\end{equation}
where $\small{\sigma_{t \to s}^2 = \Big{(} 1 / \sigma_s^2 + \alpha_{t|s}^2 / \sigma_{t|s}^2 \Big{)}^{-1}}$ and $\small{\mu_{t \to s} = \sigma_{t \to s}^2 \Big{(}\alpha_{t|s} / \sigma_{t|s}^2 \vz_t + \alpha_s / \sigma_s^2 \vx \Big{)}}$
with $\alpha_{t|s} = \alpha_t / \alpha_s$ and $\sigma_{t|s}^2 = \sigma_t^2 - \alpha_{t|s} \sigma_s^2$. 
Note that 
during sampling, the standard deviation for $p(\vz_s | \vz_t)$ is chosen to be a log-linear interpolation between $\sigma_{t|s}$ and $\sigma_{t \to s}$ of the form $\sigma_{t \to s}^{\gamma}\sigma_{t|s}^{1-\gamma}$ where $\gamma = 0.3$ was set empirically, which is important as FID is sensitive to over-smooth or grainy images.

\section{Simpler Diffusion}
\begin{figure*}
    \centering
    \begin{subfigure}[b]{0.49\textwidth}
    \centering
    \includegraphics[width=.85\textwidth]{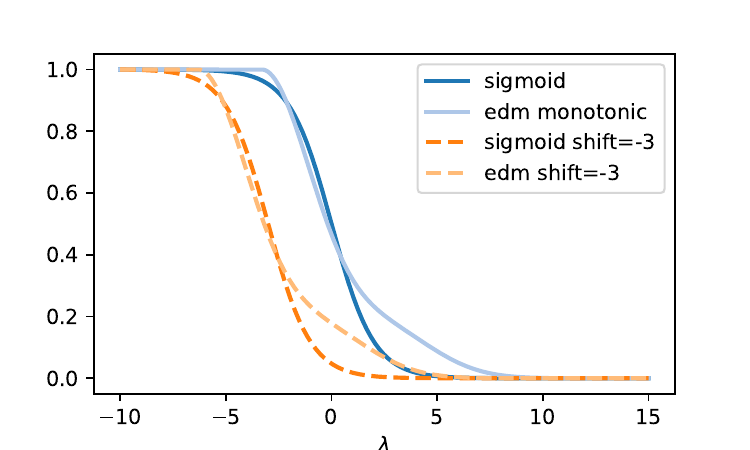}\vspace{-.1cm}
    \caption{Weighting over $\veps$-mse}
    \label{fig:eps_mse_w}
    \end{subfigure}
    \begin{subfigure}[b]{0.49\textwidth}
    \centering
    \includegraphics[width=.85\textwidth]{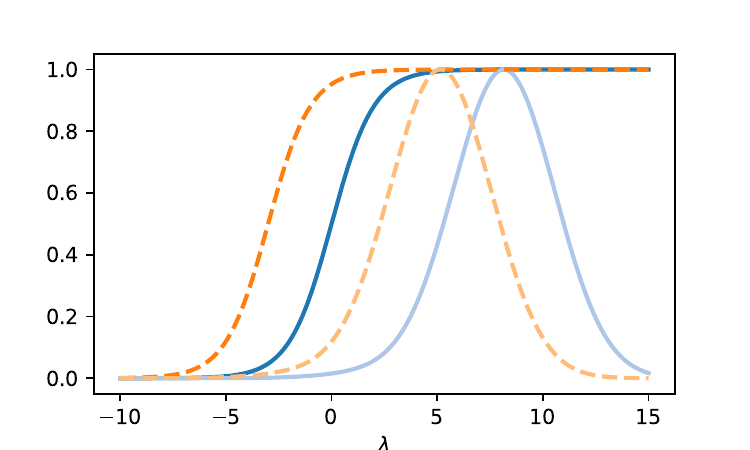}\vspace{-.1cm}
    \caption{Weighting over $\vx$-mse}
    \label{fig:x_mse_w}
    \end{subfigure}
    \caption{Weighting of different losses in both $\veps$ and $\vx$ space. Note that the weightings are scaled to have a maximum of one.}
    \label{fig:weighting_eps}
    \vspace{-.5cm}
\end{figure*}

\subsection{Revisiting the sigmoid loss}
Recall that one of the proposed losses in VDM++ \citep{kingma2023understandingdiffusion_vdmplus} is the sigmoid loss. They expressed the loss with respect to $\veps$-mse: $L(\vx) = \mathbb{E}_{t \sim \mathcal{U}(0, 1)} {-\frac{\mathrm{d}\lambda_t}{ \mathrm{d}t}} \tilde{w}(\lambda_t)  \| \veps - \hat{\veps} \|^2$, which is connected to $\vx$-mse through $\|\veps - \hat{\veps}\|^2 = \exp(\lambda_t) \| \vx - \hat{\vx} \|^2$ and $\tilde{w}(\lambda_t) = \exp(\lambda_t) w(\lambda_t)$. In VDM++, viewing the loss in $\veps$-mse enables a natural connection between the weighted loss and the ELBO and showed that losses should be monotonically decreasing on $\veps$-mse. Hence, they propose the sigmoid loss: $\tilde{w}(\lambda_t) = \sigma(b - \lambda_t)$, where $b$ denotes the bias or shift and $\sigma$ is the sigmoid function. However, they ultimately opted for another loss-weighting named EDM-monotonic for high resolution image generation, see Figure~\ref{fig:eps_mse_w}. 
\paragraph{Enforcing weighting monotonicity in $\vx$}
Revisiting the sigmoid loss, we derive its $\vx$-mse form, which turns out to conveniently be defined by a mirrored sigmoid (albeit multiplied by a constant factor):
\begin{align}
\begin{split}
    \sigma(b - \lambda_t) \| \veps - \hat{\veps} \|^2 &= \sigma(b - \lambda_t) \exp(\lambda_t) \| \vx - \hat{\vx} \|^2 \\ &= \textcolor{gray}{\exp(b)} \sigma(\lambda_t - b) \| \vx - \hat{\vx} \|^2.
\end{split}
\end{align}
This reveals that the sigmoid loss is not only monotonically decreasing in $\veps$-mse, it is also monotonically increasing in $\vx$-mse. We claim that a reasonable weighting function $w(\lambda_t)$ should be monotonically increasing over $\lambda_t$. Observe that with less noise in $\vz_t$, there is more information available and it should be strictly easier to predict the original data, meaning that the $\vx$-mse generally decreases for higher $\lambda_t$. This phenomenon is also observed in practice. If the loss \textit{weighting} in $\vx$-mse also decreases, the model would be able to make worse predictions while obtaining \textit{less noisy (better)} conditioning information and still achieve a lower loss. Simply put: Why put less weight on an easier problem?

Studying both the form of the sigmoid and EDM-monotonic weighting functions in Figure~\ref{fig:weighting_eps}, we observe that both weighting functions are monotonically decreasing in $\veps$-mse, but only the sigmoid weighting is monotonically increasing in $\vx$-mse. Therefore, we expect the sigmoid weighting to be a better candidate to scale to high resolution generation than the EDM-monotonic version.

\paragraph{Loss shifting}
Recall that for a resolution that is two times higher in both height and width, an average (low pass) filter would see $\operatorname{Var}[(x_{11} + x_{12} + x_{21} + x_{22}) / 4] = 1/4$ of the noise, while signal strength stays the same \citep{hoogeboom2023simple,kingma2023understandingdiffusion_vdmplus}. Hence, its effective logsnr is expected to increase by about $2 \log 2 \approx 1.386 \approx 1.5$ when viewed through a low-pass filter. Suppose that the optimal setting for images of 128 $\times$ 128 is a sigmoid weighting with $b \approx 0$, then one would expect the optimal setting for images of 512 $\times$ 512 to be roughly $b=-3$.\footnote{This holds \textit{under the condition that} the most important information is already contained by the $128 \times 128$ image, and this may for example not hold for images without global structure, e.g. that only contain textures.}. Empirically we find that slightly more aggressive shifts perform even better.

\subsection{Flop heavy scaling}
When one deals with a limited data budget, it can be difficult to balance scaling of models and regularization. Our small model version at $512 \times 512$ resolution starts out with 4 $\times$ 4 patching, and subsequently runs a U-ViT \citep{hoogeboom2023simple} on the 128 $\times$ 128 $\times$ 48 input (assuming the raw input is of $3$ channels). To make the model more expressive, one of the most common options is to double the number of channels for every layer, leading to $2\times$ model size. However, our experiments show that this results in diminishing returns. We hypothesize that it is because the model suffers from over-fitting when trained with limited data budget such as ImageNet. Indeed, when we train using distribution augmentation \citep{jun2020distributionaugmentation}, the best performance improves from 1.9 to 1.6. This however suggests another way to scale the model: Keep the architecture and parameter count almost the same and instead run on 2 $\times$ 2 patched input (so 256 $\times$ 256 $\times$ 12 for 512 inputs). Similar to $2\times$ model size, reducing the patching size also increases FLOPS by roughly a factor of 4. However, we found this model performs even better to about 1.5 without the need of distribution augmentation. We call this model \emph{flop heavy}, because it has a much higher flop over parameter ratio.

Flop heavy scaling has another interesting property. Suppose we have a small model trained to fit input of size $256 \times 256$ with $2 \times 2$ patching. Using the flop heavy scaling at $512 \times 512$ resolution enables us to directly finetune from the small model trained at $256 \times 256$ with exactly the same amount of parameters, whereas its counterpart version at $512 \times 512$ resolution and with $4 \times 4$ patching needs to introduce additional parameters to handle the more input channels. This method of finetuning to higher resolutions has already been popularized in the context of text-to-image generation \citep{rombach2022highresolution}.  Here we find that flop heavy scaling is a more favorable way to scale on smaller datasets such as ImageNet. Of course, in principle one could also train a model with more channels and $4 \times 4$ patching on a lower resolution, but this is not done in practice to train `small' models.

\subsection{Residual U-ViTs: Removing blockwise skip-connections}
U-Nets require that sub-sampled levels (for example via average pooling) are bypassed by skip-connections. However, this makes the architecture very different from large-scale language models such as transformers, where these skip-connections are not used and residual connections suffice. 

In this paper we use the U-ViT model of \cite{hoogeboom2023simple}, which is a variant of U-Nets with self-attention layers (with convolutional layers) being replaced by transformer blocks. 
Hence, that part of the architecture is closer to standard transformers. 

\begin{figure}
    \centering
    \includegraphics[width=\columnwidth]{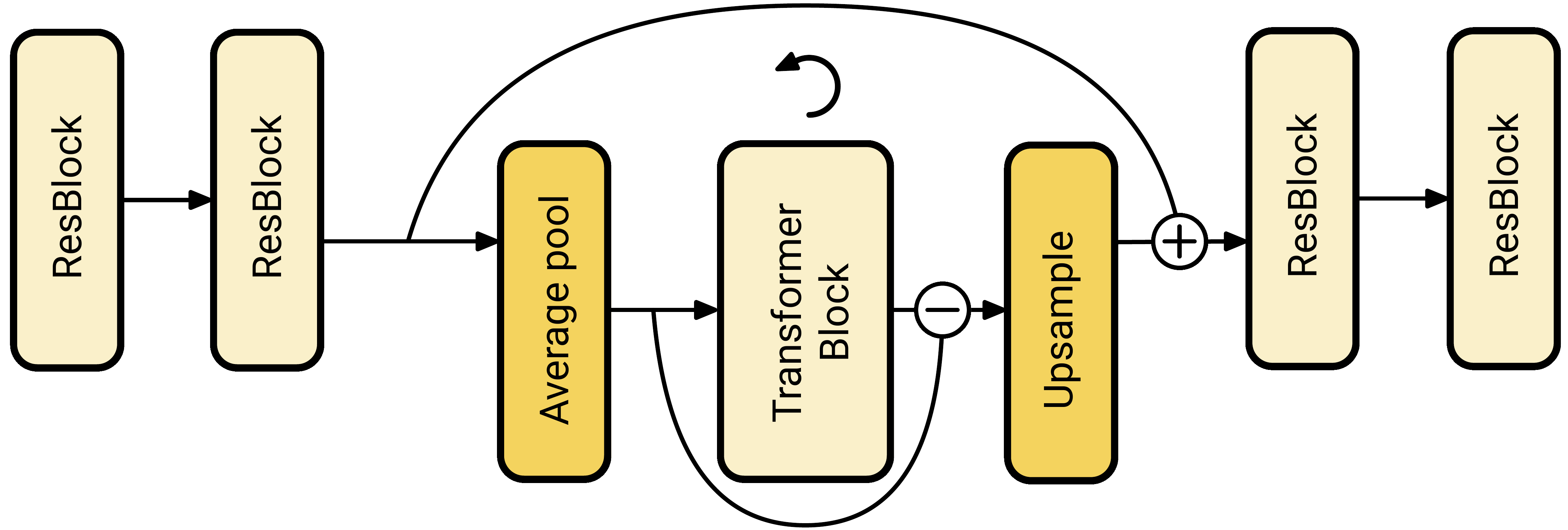}
    \caption{%
    Residual U-ViT with levelwise skip-connections (instead of blockwise).
    When the network blocks are initialized to be identity, the entire network is also an identity function at initialization.
    }
    \label{fig:residual_uvit}\vspace{-.2cm}
\end{figure}

In this paper, we further simplify model definition with Residual U-ViTs, see Figure~\ref{fig:residual_uvit}. The blockwise skip-connections are removed, and in their place only a single skip-connection is used per downsampling operation. To be precise, let $f_{d}, f_{u}$ denote a down and up level of a U-ViT and let $f_m$ denote the middle stage. This two-stage Residual U-ViT can now be described by:
\begin{equation}
    f(x) =  f_{u}\Big{(}\operatorname{U}\Big{(}f_m{(} \operatorname{D}(h)) - \operatorname{D}(h)\Big{)} + h \Big{)} \text{ where } h = f_{d}(x)
\end{equation}
where $\operatorname{D}$ is typically an average pooling and linear and $\operatorname{U}$ is typically a linear and nearest neighbour upsampling. Defining this network to preserve an identity function at initialization is now much easier: As long as $f_d$, $f_u$ and $f_m$ are initialized to be identity, the entire $f$ is also identity at initialization. The generalization to any-depth U-ViTs is also straightforward:
we can define another two levels of U-ViTs in place of $f_m$ recursively, where the nested U-ViTs can also be guaranteed to be initialized as an identity function.

For small networks, it could be helpful to keep the blockwise skip-connections, and expect to incur a slight performance degradation from removing the blockwise skip-connections. Small networks have limited capacity, so limiting skip-connections may create bottleneck issues. Experimentally, this is confirmed for the small architectures, where we observe 0.2 FID performance degradation on ImageNet512 and 0.06 FID performance degradation on ImageNet256. Because of the simplified definition, we choose to accept this performance penalty for small models. Moreover, we find that larger models perform on par or even benefit from removing the blockwise skip-connections.

The final advantage is memory consumption at evaluation. For example, a feature map at $256^2$ resolution with 128 channels requires $256^2 \cdot 128 \cdot 2 / 1024^2 = 16$MB. With 3 blocks per level, this requires holding an extra $3 \cdot 16 = 48$MB per example for the skip-connections, whereas the Residual U-ViT only requires holding $h$ and $D(h)$ (the latter could be cheaply recomputed), so $16 + 8 = 24$MB per example while computing the lower levels. For a hypothetical super-resolution model at $4096^2$ resolution, this leads to a difference between $3 \cdot 4 = 12$GB and only $4 + 2 = 6$GB.  During training this advantage is usually negligible, because feature maps also need to be retained for the backward pass. A final advantage of Residual U-ViTs is that it is more straightforward to define asymmetric U-ViTs, i.e., with a different number of blocks in the downsampling (encoding) stage from the one in the upsampling (decoding) stage. In certain cases, moving blocks from the downsampling stage to the upsampling stage can increase performance.

\subsection{Distillation}
Sampling from a pixel-space diffusion model naively often requires more function evaluations than sampling from a latent-space diffusion model. For example, the popular Stable Diffusion model~\citep{rombach2022highresolution} is usually sampled using 20-50 sampling steps, while pixel-space models are sampled with 100+ steps \citep{hoogeboom2023simple}. 
To reduce the number of sampling steps, we could use either more advanced solvers or distillation. 
Recently \emph{distillation} methods show that pixel-space models can be adapted into few-step samplers that maintain high sample quality \citep{salimans2022progressive, xiao2021tackling, kim2024consistency, yin2023one, xu2023ufogen, luo2024diff}. 
In this work, we use \emph{moment matching distillation} \citep{salimans2024multistep} 
which is currently one of the best distillation methods available in the few step regime. 
We distill our pixel-space models for use with 8-16 sampling steps, 
which empirically we find maintains or even improves the sample quality of our models. 

\figLosses

\section{Related Work}
The first end-to-end diffusion work that directly generated high resolution images was \citep{dhariwal2021diffusionbeatgans}. Given the poor performance of end-to-end models, two popular alternatives were introduced: cascaded diffusion \citep{ho2022cascaded} and latent diffusion \citep{rombach2022highresolution}.

Cascaded diffusion approaches have been shown to scale to large datasets and models to generate high quality and high resolution imagery, albeit in multiple stages \citep{ho2022cascaded, teng2024relay, zheng2024cogview3, deepfloyd2024github}, or with fine-tuning~\citep{wu2024megafusion}. Although the focus of this paper has been single stage generation up to a resolution of $1024^2$, the loss weightings may also be used for cascaded models. 

Significant improvements were made to end-to-end diffusion by shifting noise schedules \citep{hoogeboom2023simple,chen2023importancenoise} and subsequently loss weightings \citep{kingma2023understandingdiffusion_vdmplus}. Other approaches such as multi-resolution losses (and possibly architecture adaptions) \citep{gu2022fdm,gu2023matryoshka,hoogeboom2023simple} 
added marginal gains.
However, the current best known latent model EDM2 \citep{karras2023edm2} outperforms these prior end-to-end approaches by a large margin. In contrast, our models are competitive with latent models, outperforming DiT-XL/2 \citep{peebles2022scalable} and only performing slightly worse than EDM2.

\paragraph{Skip connections}
The U-Net architecture is a fully convolutional network with long-range skip-connections introduced for (medical) image segmentation~\citep{ronneberger15unet,shelhamer16fully_convolutional}. The main idea of U-Net is to combine a contracting path (which encodes the image semantics) with long skip-connections at the same resolution to enable precise localization.

With the introduction of ViTs \citep{dosovitskiy2021imageisworth} as an alternative for convolutional layers, the U-Net has been converted to U-ViTs using a similar network but replacing (some) ResNet blocks with ViTs, either with spatial down-sampling~\citep{hoogeboom2023simple} or without~\citep{bao2022all}.
Long-range skip-connections play an pivotal role in these networks, for example for fine-tuning controllable diffusion models~\citep{jiang2023scedit} and for the stability and training speed~\citep{huang2022scalelong,ma2024skiptuning}. In these latter works, it is shown that multiplying the skip-connections with a (constant) scalar ($<1$) improves training stability and speed, because of the propagation of noise throughout the network. 
In our work, we reduce the number of skip-connections significantly and initialise the network such that it resembles an identify function. 

\paragraph{Flop heavy fine-tuning} 
Fine-tuning on a higher resolution at large scale was effectively done in Stable Diffusion \citep{rombach2022highresolution} to limit the training cost, which is a form of flop heavy scaling over different resolutions. In this paper we show that flop heavy scaling is also preferred from a regularization perspective, which can be done either by finetuning from a lower resolution, or by lowering the patching size.

\figGrid

\figQualityTrainingScatter

\figScaling

\section{Experiments}
\label{sec:experiments}

\subsection{Sigmoid weighting versus others}

\paragraph{Loss weighting} In this experiment, we want to understand the effect of loss weighting on the final image quality of a model. For the noise schedule, we use an interpolated cosine schedule because it has relatively broad coverage over logsnr values. We compare the weightings corresponding to the original $\veps$-mse \citep{ho2020denoising} and the multiscale $\veps$-mse \citep{hoogeboom2023simple}, the EDM shifted weightings~\citep{karras2022elucidating}, including the original and monotonic versions \citep{karras2022elucidating, kingma2021vdm}, and the proposed sigmoid weighting with shift $b=-3$. As can be seen in Figure~\ref{fig:loss_weighting}, the sigmoid weighting outperforms all other weightings, with EDM monotonic shifted being the closest alternative. Appendix~\ref{app:additional_results} contains additional experiments, where we also evaluate the sigmoid weighting against the closest competitor without guidance. Here again sigmoid weighting is consistently better. We also evaluate a time-shifted variant, where the bias depends on the progress in training.

\paragraph{Training efficiency of noise schedules}
Since the training noise schedule can be decoupled from the loss weighting and it only affects the training efficiency but not the final performance, one can measure the convergence speed over the training process caused by leveraging particular noise schedules. In this experiment we test the cosine schedule \citep{ho2020denoising}, the shifted cosine and the interpolated cosine schedules~\citep{hoogeboom2023simple} (Figure~\ref{fig:sample_efficiency}). For fair comparison, we fix the sampling noise schedule as the cosine interpolated schedule. Interestingly, when the sigmoid weighting is used, the choice of noise schedule does not affect the training efficiency significantly. Arguably, the shifted schedule is a bit more sample efficient initially but all schedules perform similarly towards the end. This suggests that there is a lot of transfer across different logsnr values. We also tried the EDM training noise schedule with the sigma power sampling noise schedule \citep{karras2022elucidating} (not plotted), with both shifted or the original version, but its narrower logsnr sampling schedule led to performance degradation, achieving about 9 FID at best. This suggests that for high resolution, a wide logsnr range is crucial to cover the important noise levels.

\paragraph{Bias over resolutions}
In Table~\ref{tab:sigmoid_resolutions}, we show the effect of the bias term $b$ on the final performance when trained using a flop heavy model. All models use the same guidance and similar guidance intervals, where the lower bound of the guidance interval has been shifted according to resolution (again approximately 1.5 per 2x resolution change), see Appendix~\ref{app:experimental_details} for more details.

\paragraph{Sigmoid weighting versus the \textit{power loss}}
In literature, high resolution generation can greatly benefit from having a multiscale loss \citep{gu2022fdm, gu2023matryoshka,hoogeboom2023simple}. 
To that end we create a multiscale variant that we name the \textit{power loss}. 
In addition to the sigmoid weighting that effectively upweights low frequency components of the data and downweights high frequency components, the power loss aims to amplify the loss of low sub-band (smooth) signals in the spatial domain via Haar wavelets. For details on this loss see Appendix~\ref{app:power_loss}.

In Table~\ref{tab:power_loss} we see that the bias should be adjusted based on the number of Haar transform downsampling levels. Despite being competitive, the performance of power loss is roughly on par with the one with sigmoid weighting but without multiscaling (i.e. downsample levels $=0$), if we tune the bias of the sigmoid weighting more aggressively. Interestingly, lowering the bias has almost the same effect as increasing the scale. Therefore, building the multiscale variant is not necessary with the sigmoid weighting.

\subsection{Scaling channels vs tokens on ImageNet}
In this section we show the difference between scaling a model 4x FLOPS, either via channels or by creating larger resolution feature maps (or put differently, feature maps with a higher token count). Note that our small model on ImageNet512 is patched 4 x 4. This model is scaled either by increasing channels by two (named model size 2x), or by reducing patching to 2 x 2 (named flop heavy, because FLOPS over params is much higher). Indeed as can be seen in Figure~\ref{fig:scaling}, the flop heavy model performs much better than the model size 2x, when trained in a standard way on ImageNet512. Note that unlike most experiments, in these experiments it was necessary to use very different optimal guidance settings, more details in Appendix~\ref{app:experimental_details}. 

We confirm this is an issue related to overfitting, by training the model size 2x with distribution augmentation \citep{jun2020distributionaugmentation} the model size 2x variant performs much better, with an FID of 1.6 instead of 1.9 (latter in Figure~\ref{fig:scaling}). Still, without any augmentation beyond the standard horizontal flips, the flop heavy model achieves a score of 1.54.

\newcommand{\citepsmall}[1]{\small{\textcolor{gray}{\citep{#1}}}}
\begin{table}[t]
\centering
\caption{Literature Comparison on ImageNet on FID. Note that even though certain NFEs are high (e.g. 4100 for CDM) typically these suffer minor performance degradation for lower NFEs and should not be taken too literally.
} \vspace{-.3cm}
\scalebox{.99}{
\begin{tabular}{@{}lrlcl@{}}
\toprule
Method & NFE & FID  \\ \midrule
\textit{\textbf{\small Imagenet 128 x 128}} \\ \midrule
RIN \citepsmall{jabri2022scalable} & 1000 & 2.75  \\
SD, U-ViT-L \citepsmall{hoogeboom2023simple} & 512 & 1.94 \\
StyleGAN-XL \citepsmall{sauer2022styleganxl} & 1 & 1.81 \\
VDM+++ \citepsmall{kingma2023understandingdiffusion_vdmplus} & 250 & 1.75 \\
DisCo-Diff \citepsmall{xu2024discodiff} & 414 & 1.73 \\
PaGoDA \citepsmall{kim2024pagoda} & 1 & 1.48 \\ \midrule
SiD2, Flop Heavy (ours) & 512 & \textbf{1.26} \\ \midrule
\textit{\textbf{\small Imagenet 256 x 256}} \\ \midrule
ADM-G \citepsmall{dhariwal2021diffusionbeatgans} & 250 & 4.59 & \\
ADM-U \citepsmall{dhariwal2021diffusionbeatgans} & 250 & 3.94 \\
CDM \citepsmall{ho2022cascaded} & 4100 & 4.88 \\
RIN \citepsmall{jabri2022scalable,chen2023importancenoise} & 1000 & 3.52 \\
SD, U-ViT-L \citepsmall{hoogeboom2023simple} & 512 & 2.44 \\
MDM \citepsmall{gu2023matryoshka} & 250 & 3.51 \\
StyleGAN-XL \citepsmall{sauer2022styleganxl} & 1 & 2.30 \\
DiT-XL/2 \citepsmall{peebles2022scalable} & 250 & 2.27 \\
VDM++ \citepsmall{kingma2023understandingdiffusion_vdmplus} & 250 & 2.12 \\
RDM \citepsmall{teng2024relay} & 356 & 1.89\\
PaGoDA \citepsmall{kim2024pagoda} & 1 & 1.56 \\ \midrule
SiD2, small (ours) & 512 & 1.72 \\
SiD2, Flop Heavy (ours) & 512 & \textbf{1.38} \\ \midrule
\textit{\textbf{\small Imagenet 512 x 512}} \\ \midrule
ADM-G \citepsmall{dhariwal2021diffusionbeatgans} & 250 & 7.72 \\
RIN \citepsmall{jabri2022scalable,chen2023importancenoise} & 1000 & 3.95 \\
ADM-U \citepsmall{dhariwal2021diffusionbeatgans} & 250 & 3.85 \\
LDM-4-G \citepsmall{rombach2022highresolution} & 250 & 3.60 \\
SD, U-ViT-L \citepsmall{hoogeboom2023simple} & 512 & 3.02 \\
DiffiT \citepsmall{hatamizadeh2023diffit} & 250 & 2.67 \\
StyleGAN-XL \citepsmall{sauer2022styleganxl} & 1 & 2.40 \\
DiT-XL/2 \citepsmall{peebles2022scalable} & 250 & 3.04 \\
VDM++ \citepsmall{kingma2023understandingdiffusion_vdmplus} & 250 & 2.65 \\
MAGVIT-v2 \citepsmall{yu2024magvitv2} & 64 & 1.91 \\
PaGoDA \citepsmall{kim2024pagoda} & 1 & 1.80 \\
EDM2-S \small{(interval)} \citepsmall{kynkaanniemi2024guidanceinterval} & 63 & 1.68 \\
EDM2-XXL \small{(interval)} \citepsmall{kynkaanniemi2024guidanceinterval} & 63 & \textbf{1.40} \\ \midrule
SiD2, small (ours) & 512 & 2.19 \\
SiD2, Flop Heavy (ours) & 512 & 1.48 \\
SiD2, Flop Heavy distilled (ours) & 16 & 1.50\\ \midrule
\textit{\textbf{\small Imagenet 1024 x 1024}} \\ \midrule
RIN \cite{jabri2022scalable} & 1000 & 8.72 \\
SiD2, Flop Heavy (ours) & 512 & \textbf{1.75} \\
\bottomrule
\end{tabular}}\vspace{-.1cm}
\label{tab:literature}
\end{table}

\subsection{Removing block skip connection}
In this section we study the difference between the residual skip-connections and the blockwise skip-connections. As hypothesized, the small model performance drops a bit because of this. For ImageNet512 a deficit of about 0.2 FID (in Figure ~\ref{fig:skip}) and for Imagenet256 a deficit of about 0.05 FID. That the task of fitting ImageNet256 is less affected by removing skip-connections can be explained by the task being easier: The task of fitting ImageNet256 is a subset of fitting ImageNet512. 

For larger models, removing blockwise skip-connections in favor of a single residual connection does not degrade performance, and may even lead to slightly increased performance (see Figure~\ref{fig:skip}). This is somewhat unexpected, and the phenomenon persists with different guidance settings as well.  The FID curve reminds of the model size 2x curve, which at first improves but flattens out over iterations. Therefore our best hypothesis is that the block skip (flop) baseline may have similar problems such as overfitting. Regardless, it shows that Residual U-ViTs scale well.

\paragraph{Asymmetric U-ViTs}

Since our U-ViT models encompass only a single skip connection for each level, we can use a different number of ResNet or ViT blocks per level. This allows to move compute from the encoder stage (the down-sampling half of the model) to the decoder stage (the up-sampling half of the model). From Figure~\ref{fig:exp_encoder_decoder} we observe that
(a) for a fixed compute budget, it is better to add more parameters to the decoder, the encoder heavy model (E5-D1) is significantly worse than the baseline (E3-D3) and the decoder heavy model (E1-D5); (b) there needs to be some balance between the encoder and the decoder, E1-D7 works less good than E3-D5.
(c) by a slightly heavier model (E3-D7, only 10\% more parameters than our small (E3-D3) model, the performance significantly increases.

\tablemscoco

\subsection{Literature comparison}
Here we compare different models from the literature. In the first section end-to-end diffusion models are compared, and in the second section we compare between all types of generative models. In Table~\ref{tab:literature} it can be seen that our SiD2 is very competitive with existing generative models: On ImageNet128 and ImageNet256 it achieves state-of-the-art performance, on ImageNet512 it is a little worse but still competitive with the best latent diffusion model and outperforms all other models, pixel-based and latent.

\paragraph{Comparing diffusion models on compute}
The training cost SiD2 is another helpful way to see how the model scales, as depicted in Figure~\ref{figQualityTrainingScatter}). Importantly, the performance of our pixel-based approach outperforms other significantly. Observe for example the performance of our small variant (using no guidance interval to keep this comparison fair), which roughly the same performance as the VDM++ while only using a third of the training cost, and a fourth of the FLOPS in a single forward pass (see Figure~\ref{figQualityComputeScatter} in Appendix~\ref{app:additional_results}). It is also competitive with the best currently known latent diffusion approach EDM2 \citep{karras2023edm2}, but in fairness EDM2 scales a little bit better. These models attain even better performance for a smaller training cost, and a similar story holds for forward pass complexity. End-to-end diffusion can work much better than previously demonstrated, but it is still somewhat outperformed by latent diffusion on quality versus compute comparisons for both sampling and training. 

\tablekinetics

\paragraph{Text to image generation}
Finally, we show that our method can be used for end-to-end training of text-to-image models. Here, evaluation is more difficult as MSCOCO zero-shot performance depends a lot on the dataset used and is known to not necessarily correspond to actual perceived performance. Nevertheless, it is important that the model achieves performance in the right ballpark. Indeed, when we train SiD2 it performs comparable to other models in the literature on MSCOCO FID$_\text{30K}$, and detailed samples have only a small number of artifacts (Figure~\ref{fig:examples}). Furthermore, the moment matching distilled version using \citep{salimans2024multistep} outperforms the baseline model as previously observed. This model also outperforms all other zero-shot approaches at 512 resolution as shown in Table~\ref{tab:literature_tti}.

\paragraph{Video Generation}
To test whether our approach transfers to video, we test our model on Kinetics600 following \citep{ho2022videodiffusion}, by conditioning on 5 frames and generating (sometimes referred to as predicting) the subsequent 11 frames. We then compute the FVD between the training set and the generated samples both on the subsequent 11 frames. Even though our method generates at $128 \times 128$, we follow \citep{gupta2024walt} and resize samples to have resolution $64 \times 64$. Our model achieves SOTA performance and it turns out that tuning the loss is very important to achieve this as shown in Table~\ref{tab:literature_kinetics}. Note that the latent approaches (\citep{yu2022magvit, yu2024magvitv2, gupta2024walt}) use much less training compute than our models in pixel-space,
which we discuss in more detail in the Appendix~\ref{app:experimental_details}.

\section{Discussion}
In this paper we present Simpler Diffusion (SiD2) and show how one can train end-to-end diffusion and achieve competitive performance with latent models: (1) use the sigmoid loss, (2) with a simpler U-ViT architecture, and (3) favour a flop heavy model design.
As a result, our method outperforms existing end-to-end models by a large margin. 
Nevertheless, at this moment latent diffusion methods still seem to have slightly better scaling properties and it remains to be seen whether this gap can be closed any further. Importantly, for many applications training a separate autoencoder may not be either practical or desired. Therefore, we believe this paper is an important contribution that allows end-to-end modelling directly in the pixel-space.

\clearpage

{
    \small
    \bibliographystyle{ieeenat_fullname}
    \bibliography{main}
}

\clearpage
\appendix

\clearpage
\setcounter{page}{1}
\maketitlesupplementary

\section{Connection between sigmoid loss and low-bit training}
Weighted diffusion losses can be closely related to low-bit training, as shown in \cite{kingma2023understandingdiffusion_vdmplus}. In earlier normalizing flow work, low-bit training has been shown to achieve higher perceptual quality \cite{kingma2018glow}. The reason that low-bit training improves sample fidelity is quite intuitive: by throwing away less significant (and thus important) bits from the training data, the model spends more capacity on modeling more significant bits. Here we will show how sigmoid loss is related to low-bit training, and therefore provide a theoretical understanding on how sigmoid loss balances the bits from the data of different importance. 

Training diffusion models on different bit precisions and their loss contributions may give results that are entangled with optimization hyperparameters. We instead consider a simplified data distribution setting: assume the data is univariate, with a uniform distribution over the $2^n$ possible values, where n is the bit precision. We further assume the model to be optimal for all noise levels, which is a mixture-of-Gaussians whose means are the $2^n$ possible values and variance is determined by the log signal-to-noise ratio $\lambda$. A diffusion loss over the entire dataset can be expressed as: 
\begin{align}
    L &= \mathbb{E}_{q(\vx)} \mathbb{E}_{\lambda \sim q(\lambda)} w(\lambda)  \| \vx - \hat{\vx}_\lambda \|^2,\\
    & = \mathbb{E}_{q(\vx)} \mathbb{E}_{\lambda \sim q(\lambda)} \tilde{w}(\lambda)  \| \veps - \hat{\veps}_\lambda \|^2,\\
    &= \mathbb{E}_{\lambda \sim q(\lambda)} \tilde{w}(\lambda) \mathbb{E}_{q(\vx)} \| \veps - \hat{\veps}_\lambda \|^2,
    \label{eq:loss_appendix}
\end{align}
where $q(\lambda)$ is determined by the mapping $\lambda_t$ and $t\sim \mathcal{U}(0, 1)$, and $q(\vx)$ is the data distribution. Given the simplified data distribution and the optimal model, we can compute $\mathbb{E}_{q(\vx)} \| \veps - \hat{\veps}_\lambda \|^2$ analytically. Figure~\ref{fig:deps} shows its value over $\lambda$ for multiple data distributions of different bit precision $n$.

\figLowBitLoss

We can also plot the weighted loss with weighting function $\tilde{w}(\lambda)$ at different noise levels $\tilde{w}(\lambda) \mathbb{E}_{q(\vx)} \| \veps - \hat{\veps}_\lambda \|^2$. Given that the optimal sigmoid weighting at resolution $128$ is $\tilde{w}(\lambda) = \sigmoid(1 - \lambda)$, we follow \citep{hoogeboom2023simple} to shift the weighting to $\tilde{w}(\lambda) = \sigmoid(10.7 - \lambda)$ for the univariate example, where the additional bias term comes from $2 \log(128)$. The weighted loss is shown in Figure~\ref{fig:mix_loss}. Interestingly, if we follow the low-bit training but instead of assuming the data is a uniform distribution at a certain precision, we assume it is a mixture of multiple precisions, and then the loss ends up being very similar to the one with the sigmoid weighting (Figure~\ref{fig:mix_loss}). The mixture we visualize here is $[8, 7, 6, 5]$-bit with a mixture proportion of $[1, 4, 4, 6]$. Informally speaking, the loss of the sigmoid weighting is very similar training on a lower-bit distribution of the data, where the data is only modelled at full precision occasionally. 

\figLowBitWeighted

\section{Experimental Details}
\label{app:experimental_details}

\subsection{Sigmoid Loss}

The sigmoid loss can be easily implemented in the following way:
\label{app:sigmoid_loss}
\begin{lstlisting}[style=python]
def sigmoid_loss(x, model_x, logsnr_fn, t, bias):
    logsnr = logsnr_fn(t)
    dlogsnr_dt = jax.jvp(
        logsnr_fn, t (ones_like(t),))[1]
    weight = -0.5 * dlogsnr_dt * exp(
          bias) * sigmoid(logsnr - bias)
    return weight * mean_except_batch((x-model_x)**2)
\end{lstlisting}

\noindent
For completeness, recall that a cosine interpolated logsnr schedule function from \citep{hoogeboom2023simple} can be defined as:
\begin{lstlisting}[style=python]
def cosine_interpolated(
    t, lognsr_min=-10, logsnr_max=10,
    image_res=512, noise_res_low=32,
    noise_res_high=512)
  log_change_high = log(image_res) - log(noise_res_high)
  log_change_low = log(image_res) - log(noise_res_low)

  b = arctan(exp(-0.5 * logsnr_max))
  a = arctan(exp(-0.5 * logsnr_min)) - b
  logsnr_cosine = -2. * (jnp.log(jnp.tan(a * t + b))
  logsnr_high = logsnr_cosine + log_change_high
  logsnr_low = logsnr_cosine + log_change_low
  return (1 - t) * logsnr_high + t * logsnr_low
\end{lstlisting}
\noindent Observe here that the min and max values are shifted along as well and are not the limits of the resulting function.

\subsection{Power loss}
\label{app:power_loss}
Recall that the power loss aims to amplify the loss of low sub-band (smooth) signals in the spatial domain via Haar wavelets. \textit{Averaging (ie, low passes) increases the logsnr, therefore we adjust the logsnr based on the number of low passes}. We apply the Haar wavelet transformation iteratively to the input $I = \hat{\vx} - \vx$. The Haar wavelet $W(I) = (L, H)$ transforms the input into a low sub-band (smooth) $s_k = (i_{2k} + i_{2k + 1}) / \sqrt{2}$ and a high sub-band (detail) $d_k = (i_{2k} - i_{2k + 1}) / \sqrt{2}$. First $W$ is applied to the rows $W_r(I) = (L, H)$ and then to the columns on each sub-band $W_c(L) = (LL, LH)$ and $W_c(H) = (HL, HH)$. To give an example, for a $512^2$ image, applying this two times results in the sub-bands $S = \{LLLL, LLLH, LLHL, LLHH, LH, HL, HH\}$ where the first four sub-bands have a resolution of $128^2$ and the last three have a resolution of $256^2$.

Intuitively we want to emphasize high frequencies at the higher resolutions and low frequencies at the lower resolutions. Therefore we shift the logsnr per sub-band roughly according to the expected increase in signal-to-noise. We calculate the shift for a sub-band as $b_s = \log(2) \cdot l(s) - b$ where $l(s)$ is the number of low passes on the sub-band $s$ (ie, the number of $L$'s in the sub-band). Just like the sigmoid loss we cancel out the effect of weighting of the noise schedule on the loss $w_s = \sigmoid(\lambda_t + b_s) \cdot -\frac{\mathrm{d}\lambda_t}{\mathrm{d}t}$. Lastly we sum over each sub-band and multiply the sub-band weight with the squared sum of the sub-band $PL(I) = \sum_{s \in S} w_s(\lambda_t) ||s||^2$.

\subsection{Estimating Flops}
Although FLOPs can be estimated directly using accelerator tooling, it turns out that the most significant operations outweigh all other minor operations (within 1\%). These most significant operations are matrix multiplications (including convolutional layers) and the self-attention dot product. The compute footprint of these operations for our architecture can be calculated using the following equations. Consistent with \citep{karras2023edm2}, we assume a training step has the computational cost of a forward pass times 3, and that multiply-adds are counted as \textit{one} flop.

\begin{lstlisting}[style=python]
def transformer_gflops(size, num_channels, blocks):
    # q, k, v, attn_out, mlp in (4), mlp out (4).
    linears = 12 * num_channels ** 2 * blocks * size ** 2
    attn = 2 * size ** 4 * blocks * num_channels
    return (linears + attn) / 1000**3

def resblock_gflops(size, num_channels, blocks):
  flops = 2 * 3**2 * blocks  # 2 layers with 3x3 conv
  flops *= num_channels ** 2 # channels
  flops *= size ** 2  # spatial resolution
  return flops / 1000**3
\end{lstlisting}

\subsection{Guidance Intervals}
We use guidance intervals \cite{kynkaanniemi2024guidanceinterval}, in which classifier-free guidance is only applied for noise levels within the interval and disabled elsewhere. Based on a grid search we found that the combination of guidance 1.0 on logsnr (-3, +5) obtained the best FID quality on ImageNet512. 
Without further tuning we found that the minimum logsnr could simply be shifted by approximately 1.5 for a 2x resolution change, similar to how loss biases are shifted. 
The logsnr max value is typically less sensitive~\cite{kynkaanniemi2024guidanceinterval}, and was kept constant.

\subsection{A note on auto-guidance}
Besides applying guidance on intervals \citep{kynkaanniemi2024guidanceinterval}, there is another technique named autoguidance \citep{karras2024autoguidance} which can even produce better FIDs with EDM-XXL (1.2 from 1.4 with guidance intervals). Autoguidance does not use an unconditional model as negative signal. Instead, the negative signal is provided by a \textit{worse} model, either earlier in training or a small version. Autoguidance increases the hyperparameter space of guidance even further, and we found that this made autoguidance more difficult to tune. For that reason we opted to compare all methods in literature on guidance intervals if available, and also provided the performance of our model with constant guidance to compare with older methods.  

\subsection{Experimental settings}
The small model variant uses the following settings. 
\begin{lstlisting}[style=python]
channels = [128, 256, 512, 1024]
num_updown_blocks = [3, 3, 3],
num_mid_blocks = 16,
block_dropout = [0., 0., 0.1, 0.1],
block_type = ['ResBlock', 'ResBlock',
              'Transformer', 'Transformer'],
mean_type = v
loss_type = sigmoid:-3  # 512^2
patching_size = 4
loss_type = sigmoid:-1  # 256^2
patching_size = 2
loss_type = sigmoid:0   # 128^2
patching_size = 1
\end{lstlisting}

The flop heavy variant uses the following settings:
\begin{lstlisting}[style=python]
channels = [128, 256, 512, 1024]
num_updown_blocks = [3, 3, 3]
num_mid_blocks = 16
block_dropout = [0., 0., 0.1, 0.1]
block_type = ['ResBlock', 'ResBlock',
              'Transformer', 'Transformer']
mean_type = v
loss_type = sigmoid:-3  # 512^2
patching_size = 2
loss_type = sigmoid:-1  # 256^2
patching_size = 1

loss_type = sigmoid:0   # 128^2
patching_size = 1
# for 128^2 resolutions the top-level
# layer (128 channels) is entirely removed.
\end{lstlisting}

To our surprise, a grid search over guidance strength and guidance interval determined that the same settings were optimal for the small and flop heavy variant on ImageNet512. We then shifted the lower bound of the guidance interval based on resolution shift. Resulting in: 
\begin{lstlisting}[style=python]
guidance_interval = (-3, 5)    # 512^2
guidance_interval = (-1.5, 5)  # 256^2
guidance_interval = (0., 5)    # 128^2
guidance = 1.0
num_steps = 512
sampler = 'ddpm'
clip_x = 'static'
logvar_type = '0.3'
\end{lstlisting}

The training settings are:
\begin{lstlisting}[style=python]
batch_size=2048
optimizer='adam'
adam_beta1 = 0.9
adam_beta2 = 0.99
adam_eps = 1.e-12
diffusion_schedule =
    'cosine_interpolated_low_32_high_512'
learning_rate=1e-4
learning_rate_warmup_steps=10_000
weight_decay=0.0
ema_decay=0.9999
max_train_steps = 1_000_000  # for small variants
max_train_steps = 800_000    # for flop heavy
\end{lstlisting}

\subsection{Kinetics600}
The training settings are:
\begin{lstlisting}[style=python]
batch_size=512
optimizer='adam',
adam_beta1=0.9
adam_beta2=0.98
adam_eps=1.e-10
learning_rate=1e-4
learning_rate_warmup_steps=100
diffusion_schedule = 
    'cosine_interpolated_low_32_high_128'
weight_decay=0.0
ema_decay=0.9999
max_train_steps = 500_000
\end{lstlisting}

\noindent And the model settings are:
\begin{lstlisting}[style=python]
channels = [128, 256, 512, 1024]
num_updown_blocks = [3, 3, 4],
num_mid_blocks = 8,
block_dropout = [0., 0., 0.1, 0.1],
block_type = ['ResBlockConv3D', 'ResBlockConv3D',
              'LocalTransformer_19_19_19',
              'Transformer'],
mean_type = v
loss_type = sigmoid:0
guidance = 0.  # unconditional
num_steps = 128
sampler = 'aDDIM'
sampler_noise = 'data_0.9'
clip_x = 'static'
\end{lstlisting}

\noindent Using the same gflop calculations as before adapted to 3D Convs, our 3570 GFLOPs per forward pass, and about 2.5 zettaflops for training. For reference, the base model of W.A.L.T. uses approximately 0.5 zettaflops (not counting the autoencoder training). This shows that pixel-based models can outperform latent approaches and scale much better using our tuning than before. On the other hand, they still require more compute to achieve this performance that latent approaches. During sampling we use aDDIM \citep{heek2024multistep} because we found that it allowed fewer sampling steps ($128$) with acceptable performance without much noise schedule tuning, making evaluation during training a bit faster.

\figNoGuidanceLoss

\figTimeShiftedSigmoidAppendix

\figEncoderDecoderAppendix

\figQualityComputeScatter

\begin{table*}
\centering
\caption{Comparison on sampling cost in FLOPs and generation methods.
}
\newcommand{\mult}[2]{#1 * #2}
\label{tab:sampling_comparison}
\scalebox{1.0}{
\begin{tabular}{@{}lrrrrrrr@{}}
\toprule
Method & NFE & Inference cost (GFLOPs) & Sampling cost (TFLOPs) & Sampling time (s) & FID \\ \midrule
StyleGAN-XL \citepsmall{sauer2022styleganxl} & 1 & - & - & 0.10 & 2.40 \\
EDM2-S \citep{karras2023edm2} & 63 & \rawdata{img512-GI-S-gflops0} & 42.5 & - & 1.68 \\
EDM2-XXL \citep{karras2023edm2} & 63 & \rawdata{img512-GI-XXL-gflops0} & 61.6 & - & \textbf{1.40} \\
DiT-XL/2 \citep{peebles2022scalable} & 250 & \rawdata{img512-DiT-XL-gflops0} & 262.5 & - & \rawdata{img512-DiT-XL-cfg2} \\
Pagoda \citep{kim2024pagoda} & 1 & - & - & \textbf{0.05} & 1.80 \\
\midrule
SiD2 16-step distilled (ours) & 16 & \rawdata{img512-SD2-flop-gflops0} & 10.4 & 0.29 & 1.50 \\
\bottomrule
\end{tabular}}
\end{table*}

\section{Additional Results}
\label{app:additional_results}
\subsection{No Guidance}
To validate the difference between the sigmoid and edm-monotonic loss even further, we analyze the FID performance over training iterations without applying any guidance to either model. As can be seen in Figure~\ref{fig:noguidance}, the difference in performance is even more pronounced when no guidance is applied.

\subsection{Timeshifted Sigmoid Losses}
\label{app:timeshift_sigmoid}
The bias term in the sigmoid loss function could be used to emphasize on the low-level information (bits) of the generated images.
One could argue that during the course of training, it is important to first focus on the structure of the image and only later penalise mismatches in the high-frequency details. 
Therefore, we ran a set of experiments with a time shifted bias, where the bias is increased with a linear warmup, we start with bias $b_{\textrm{start}}$ and interpolate over $t_{b}$ steps towards $b_{\textrm{end}}$, yielding: 
$b_t = b_{\textrm{start}} + (b_{\textrm{end}}-b_{\textrm{start}}) \cdot \min(t / t_{\textrm{b}}, 1)$. 
So the the bias is interpolated from a starting value to the (fixed) end value (-3) in 100k or 200k training steps.

In Figure~\ref{fig:exp_timeshift_all}, we report the FID over the course of training. 
From the results we observe that shifting the loss during training generally leads to a small improvement, albeit resulting in two additional hyper parameters (starting value and number of steps). 
Especially when the bias is not too low: shifting from bias -8/-6 to -3 is a good idea, as long as the warmup is relatively quick (100k iterations). Surprisingly, for warmups to 200k iterations the time shifted models have worse performance in early training of stages.

\subsection{Asymmetric U-ViTs}
\label{app:asymmetric_u_vit}
In our U-ViT design only a single skip connection per encoding/decoding level is used, allowing for different number of blocks in each level.
Here we experiment using different numbers of encoding blocks versus decoding blocks at the same level.
The results are in Figure~\ref{fig:exp_asymmetric_uvit}. 
We see that increasing depth helps, and decoder could be a little heavier than the encoder, although symmetric scaling (E3-D3, our baseline) already acts as a strong baseline.

\subsection{Forward pass complexity}
In addition to the training cost in the main text, Figure~\ref{figQualityComputeScatter} shows the forward pass complexity. Here we see that again our method outperforms all existing pixel approaches by a large margin, and performs similar albeit somewhat worse than latent approaches.

\subsection{Additional samples}
\begin{figure*}
    \centering
    \includegraphics[width=.12\textwidth]{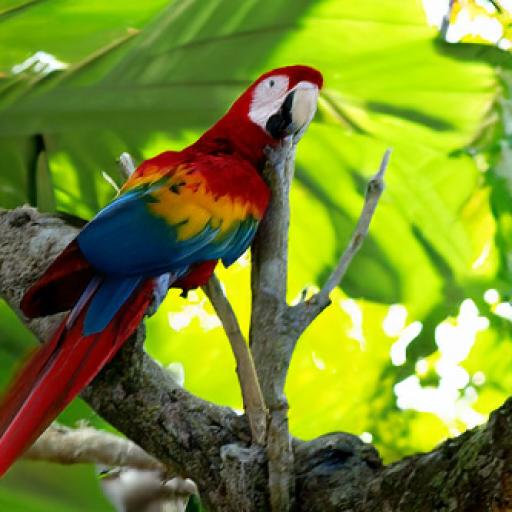} \hfill
    \includegraphics[width=.12\textwidth]{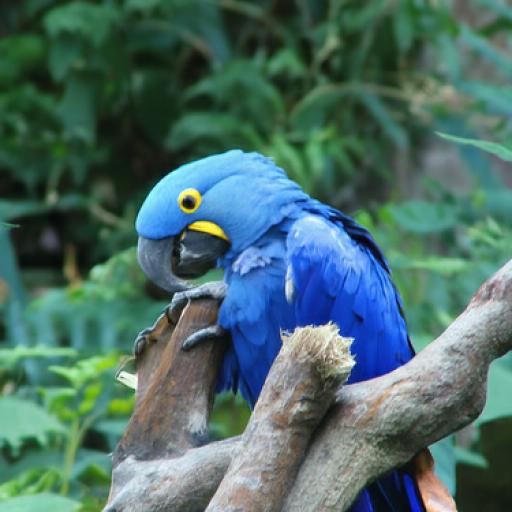} \hfill
    \includegraphics[width=.12\textwidth]{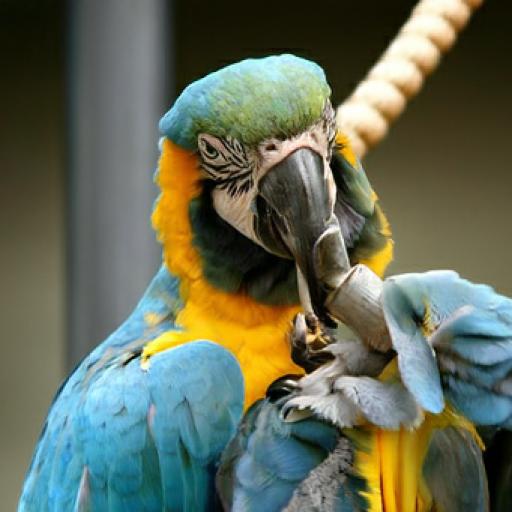} \hfill
    \includegraphics[width=.12\textwidth]{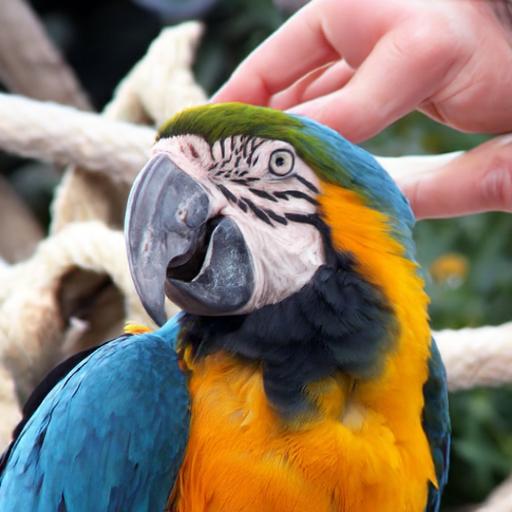} \hfill
    \includegraphics[width=.12\textwidth]{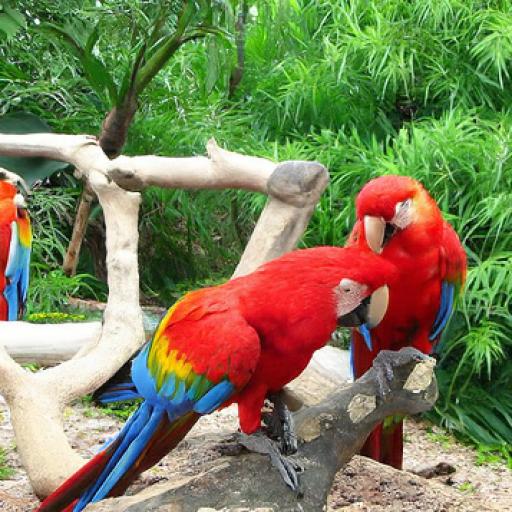} \hfill
    \includegraphics[width=.12\textwidth]{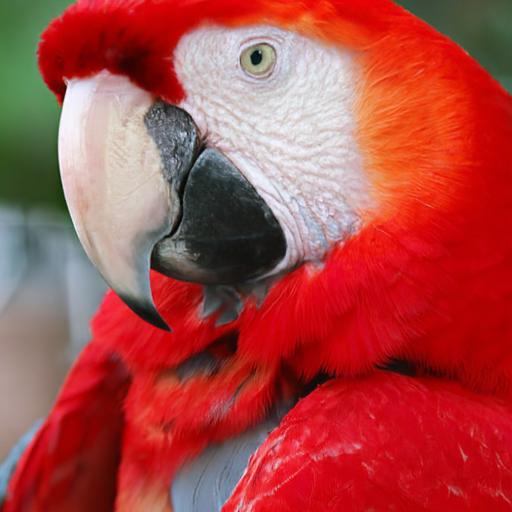} \hfill
    \includegraphics[width=.12\textwidth]{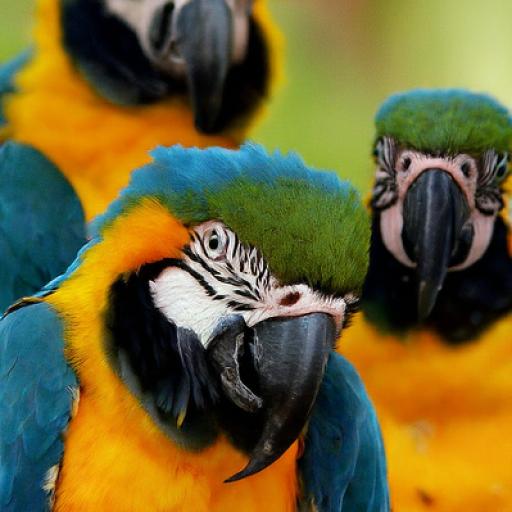} \hfill
    \includegraphics[width=.12\textwidth]{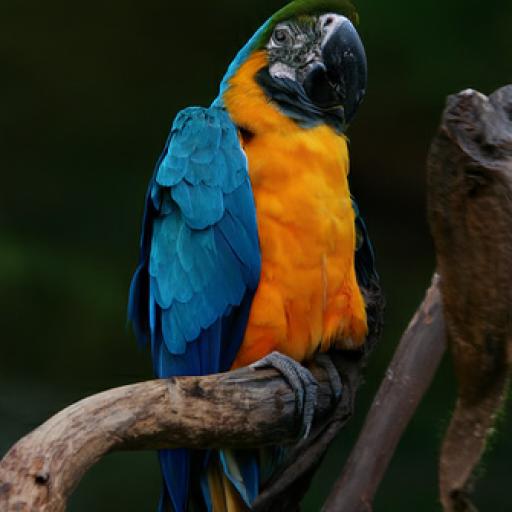} \hfill
    \includegraphics[width=.12\textwidth]{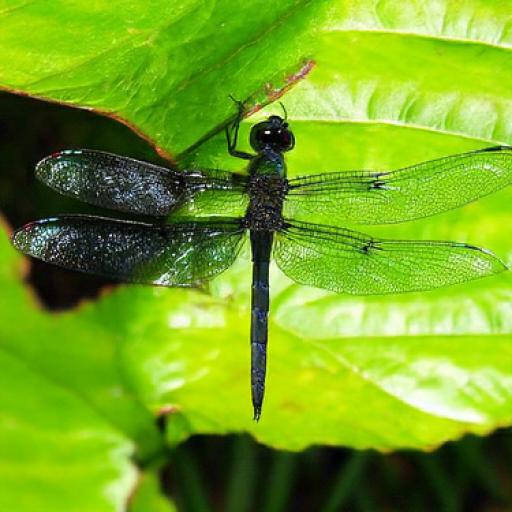} \hfill
    \includegraphics[width=.12\textwidth]{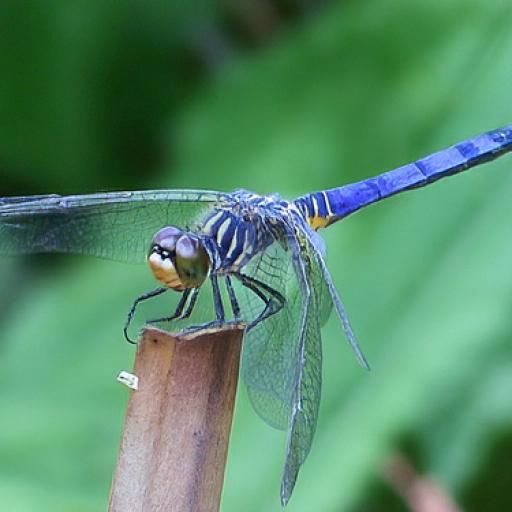} \hfill
    \includegraphics[width=.12\textwidth]{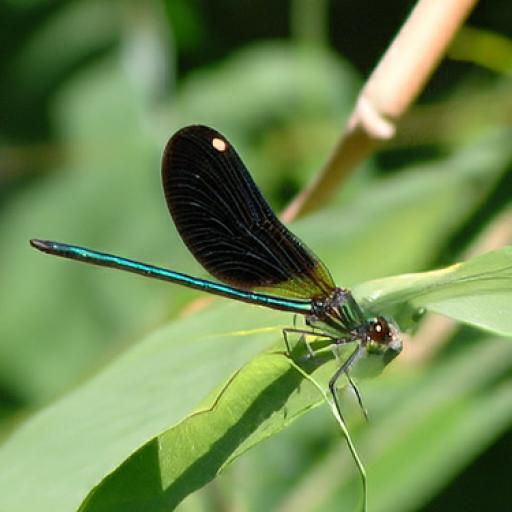} \hfill
    \includegraphics[width=.12\textwidth]{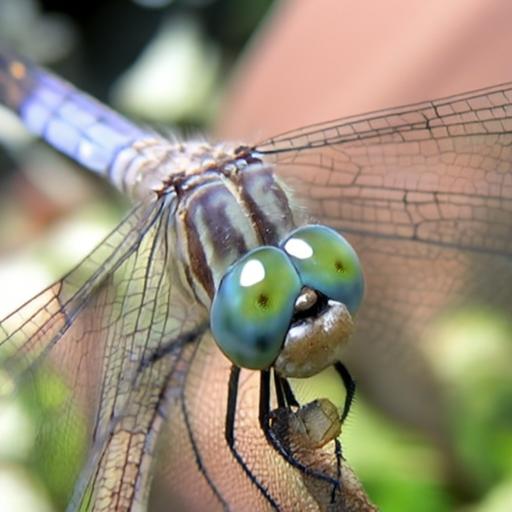} \hfill
    \includegraphics[width=.12\textwidth]{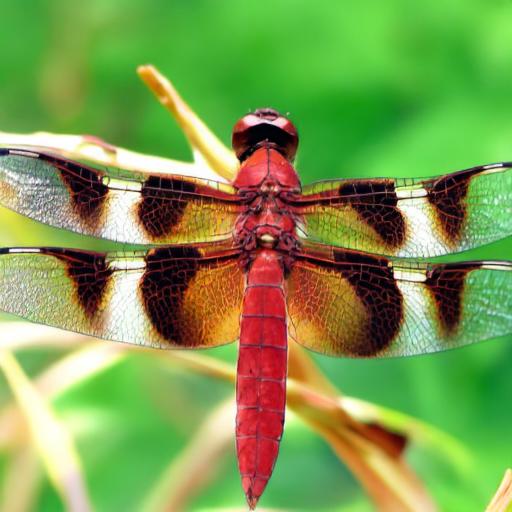} \hfill
    \includegraphics[width=.12\textwidth]{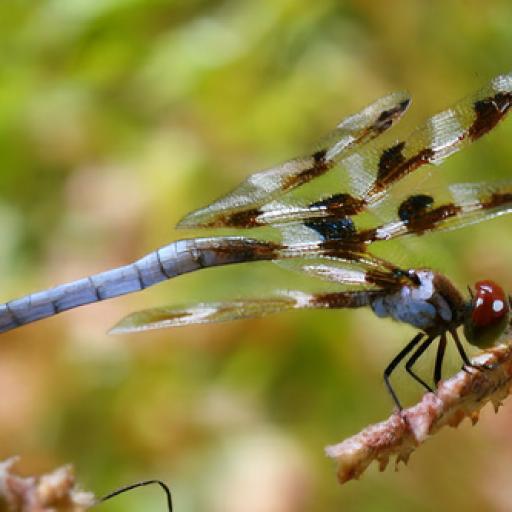} \hfill
    \includegraphics[width=.12\textwidth]{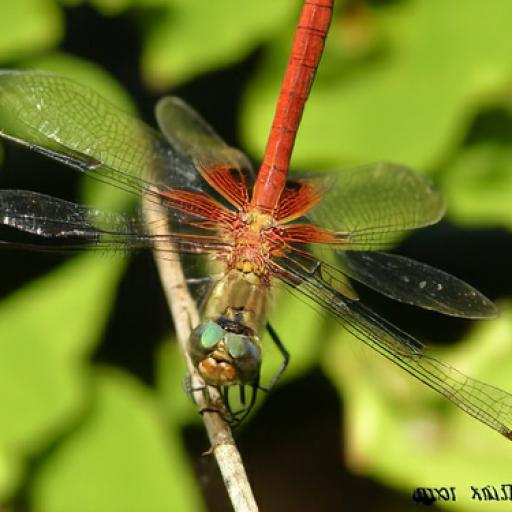} \hfill
    \includegraphics[width=.12\textwidth]{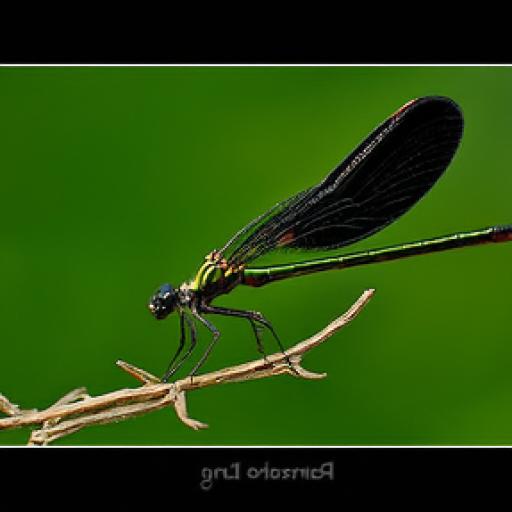} \hfill
    \includegraphics[width=.12\textwidth]{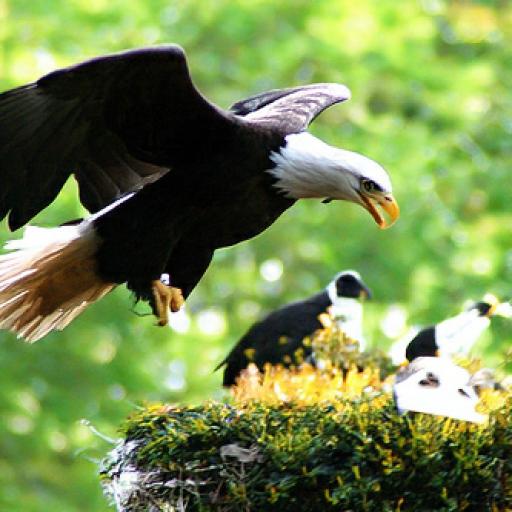} \hfill
    \includegraphics[width=.12\textwidth]{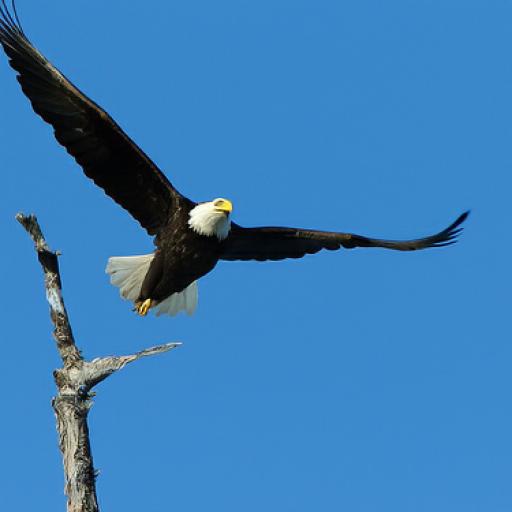} \hfill
    \includegraphics[width=.12\textwidth]{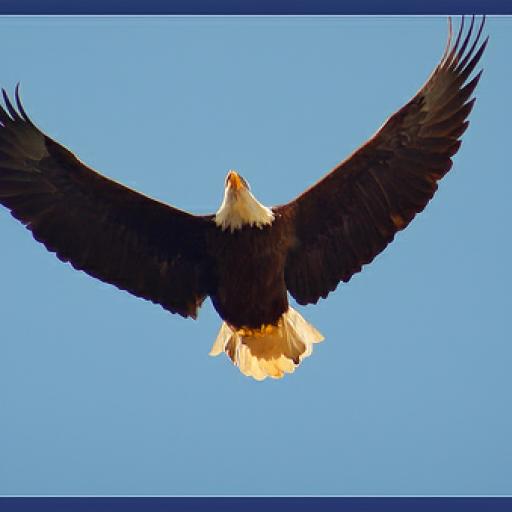} \hfill
    \includegraphics[width=.12\textwidth]{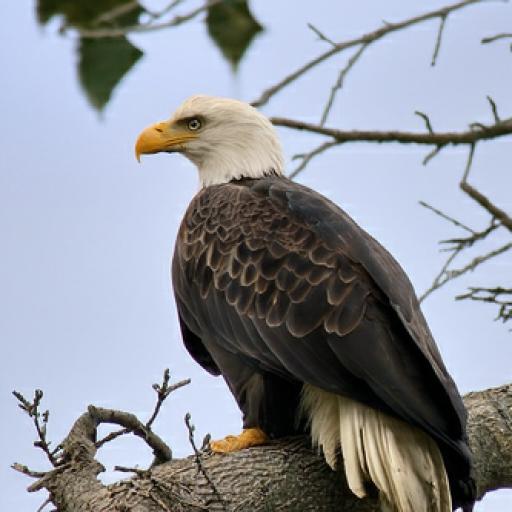} \hfill
    \includegraphics[width=.12\textwidth]{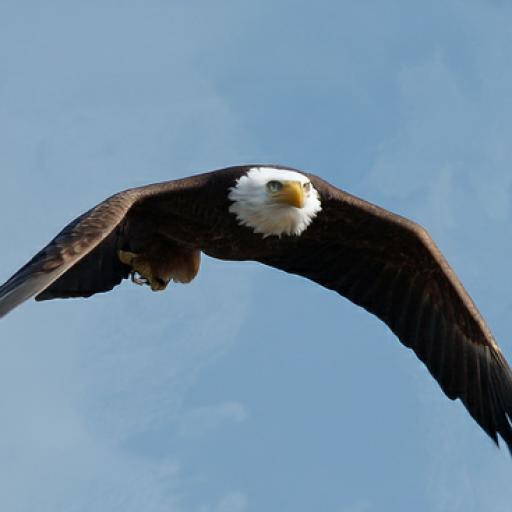} \hfill
    \includegraphics[width=.12\textwidth]{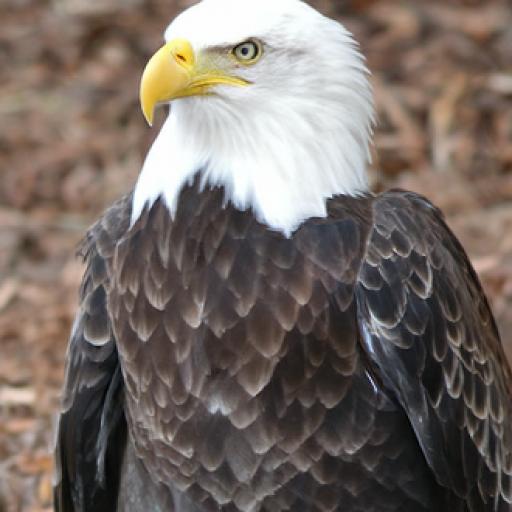} \hfill
    \includegraphics[width=.12\textwidth]{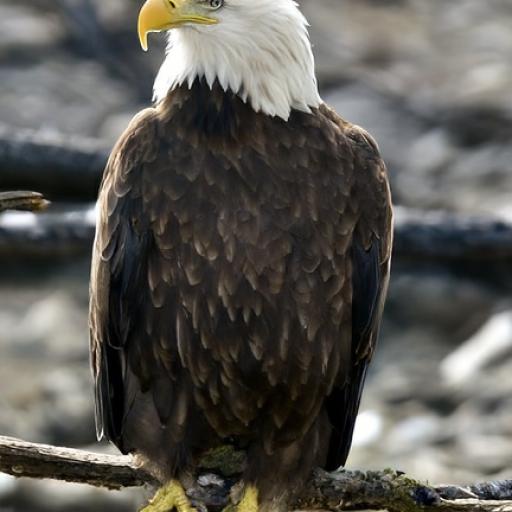} \hfill
    \includegraphics[width=.12\textwidth]{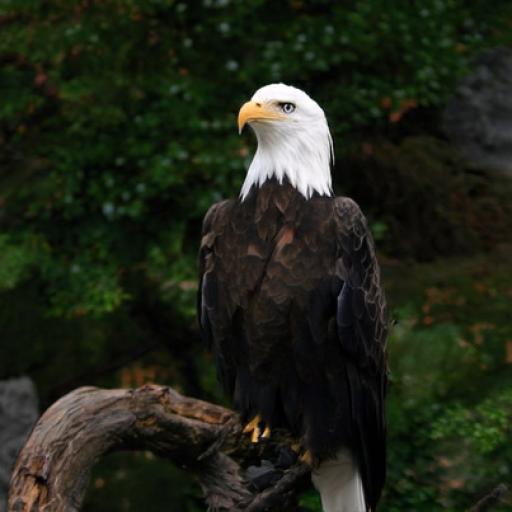} \hfill
    \includegraphics[width=.12\textwidth]{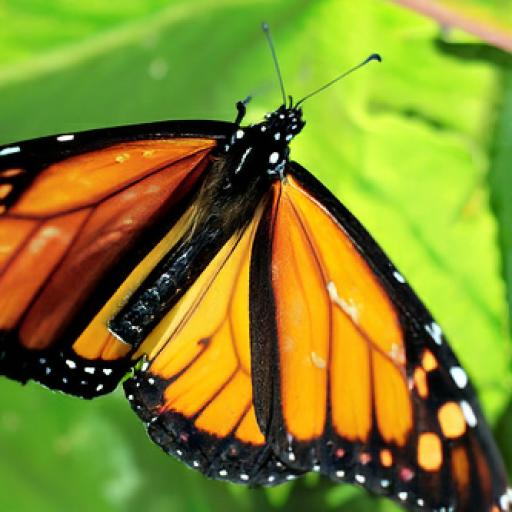} \hfill
    \includegraphics[width=.12\textwidth]{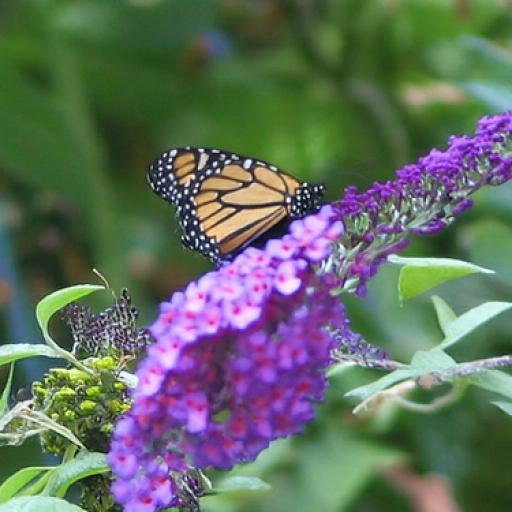} \hfill
    \includegraphics[width=.12\textwidth]{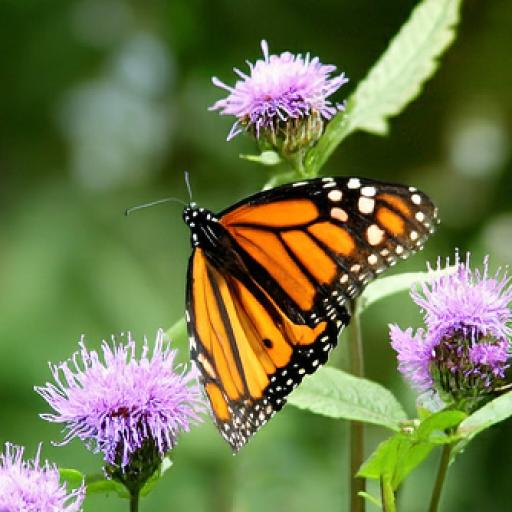} \hfill
    \includegraphics[width=.12\textwidth]{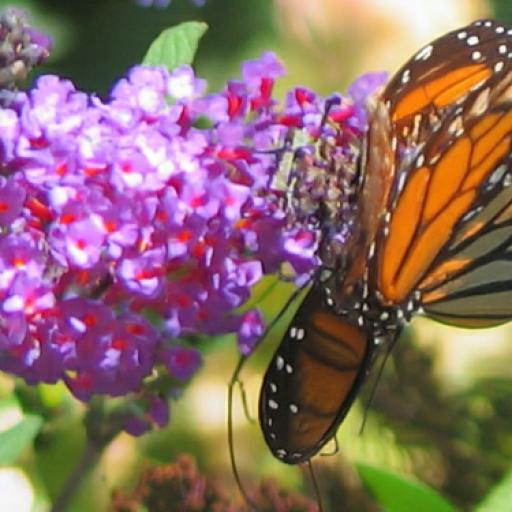} \hfill
    \includegraphics[width=.12\textwidth]{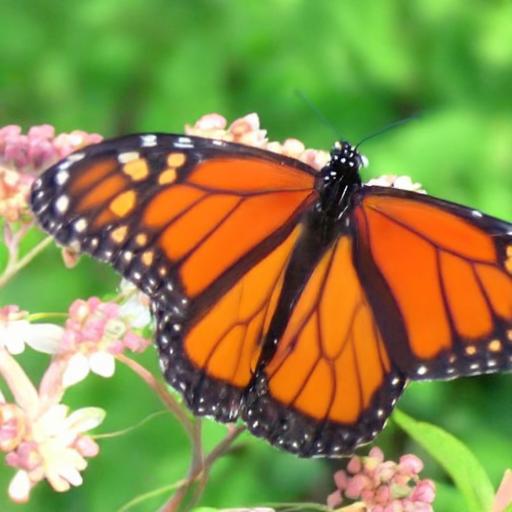} \hfill
    \includegraphics[width=.12\textwidth]{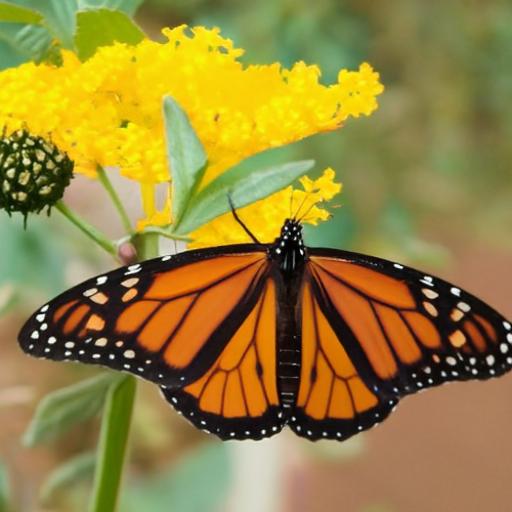} \hfill
    \includegraphics[width=.12\textwidth]{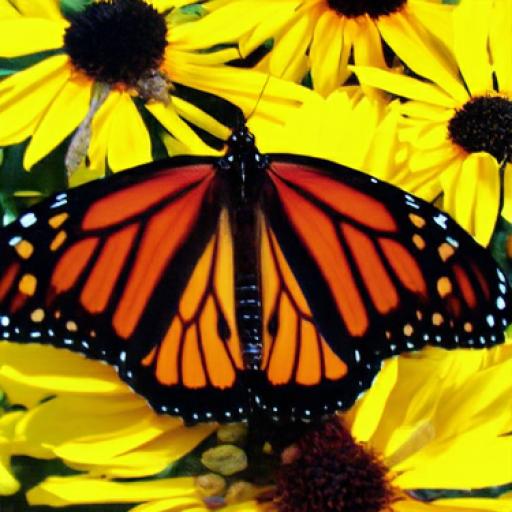} \hfill
    \includegraphics[width=.12\textwidth]{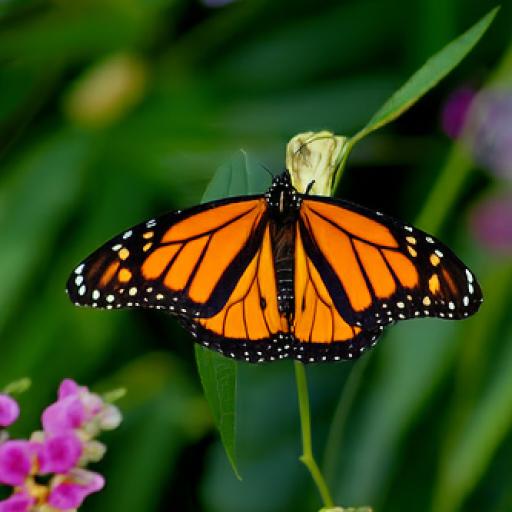} \hfill
    \includegraphics[width=.12\textwidth]{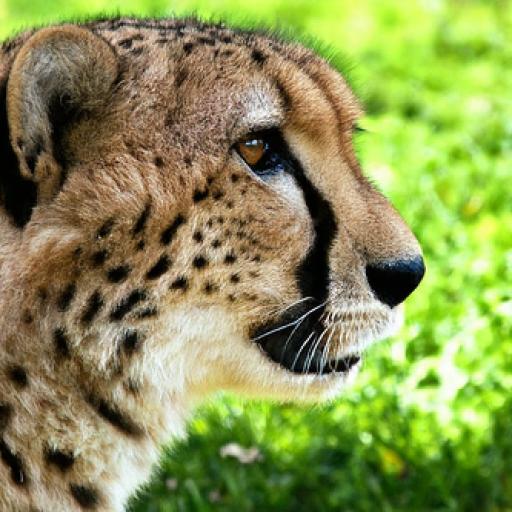} \hfill
    \includegraphics[width=.12\textwidth]{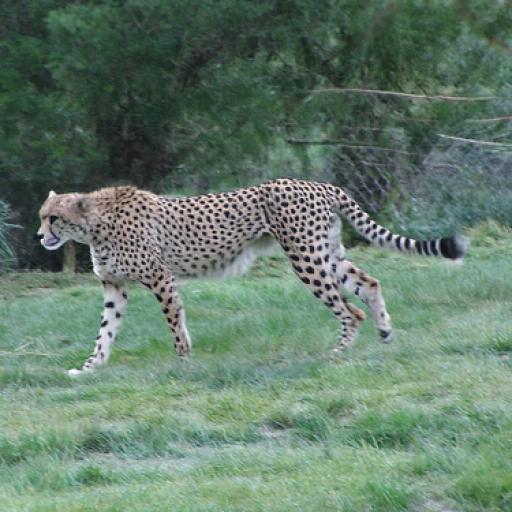} \hfill
    \includegraphics[width=.12\textwidth]{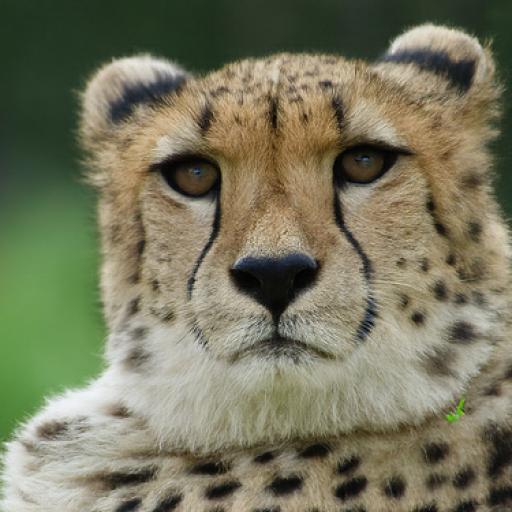} \hfill
    \includegraphics[width=.12\textwidth]{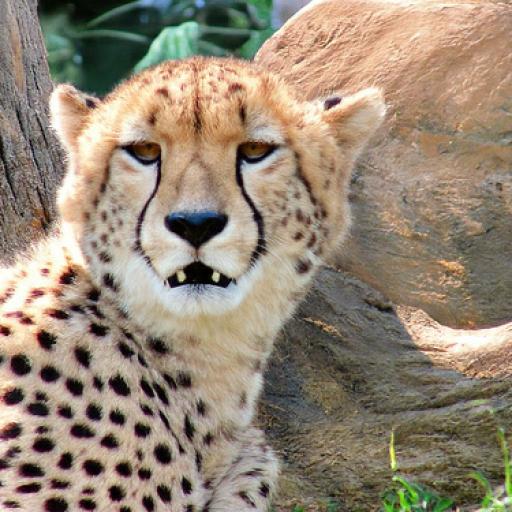} \hfill
    \includegraphics[width=.12\textwidth]{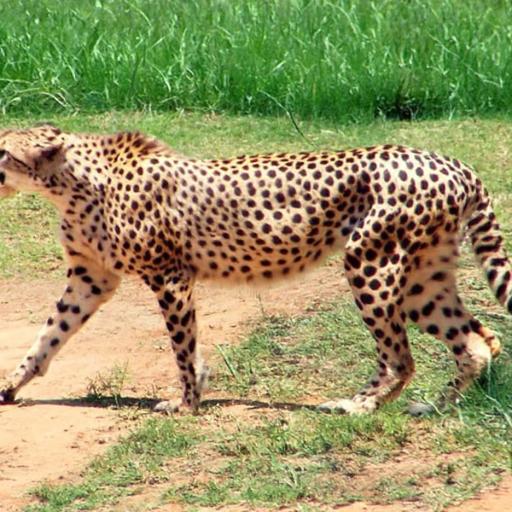} \hfill
    \includegraphics[width=.12\textwidth]{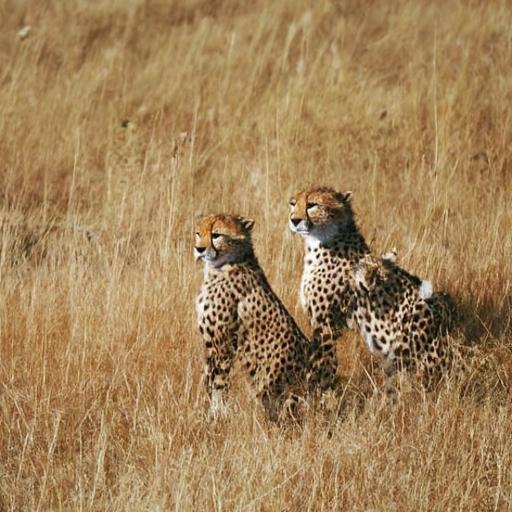} \hfill
    \includegraphics[width=.12\textwidth]{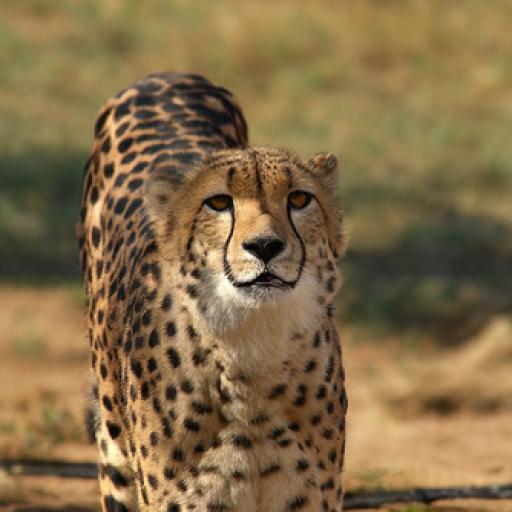} \hfill
    \includegraphics[width=.12\textwidth]{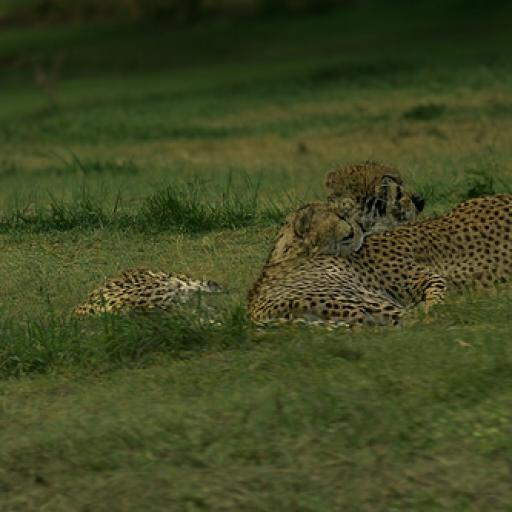} \hfill
    \includegraphics[width=.12\textwidth]{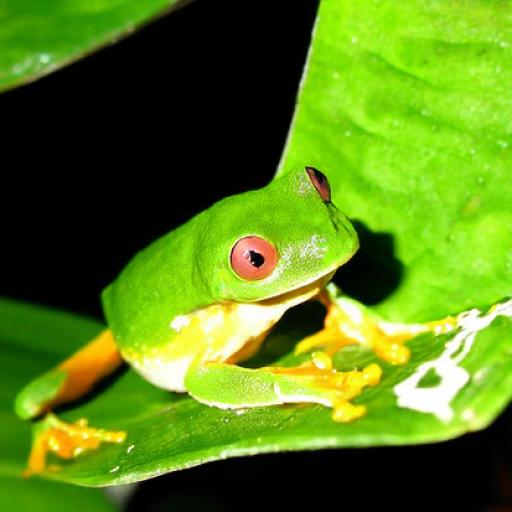} \hfill
    \includegraphics[width=.12\textwidth]{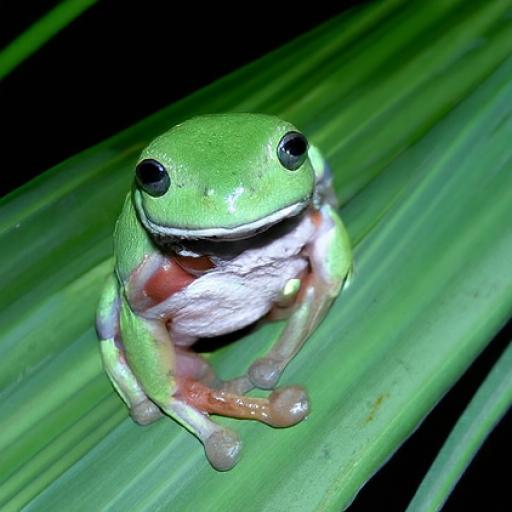} \hfill
    \includegraphics[width=.12\textwidth]{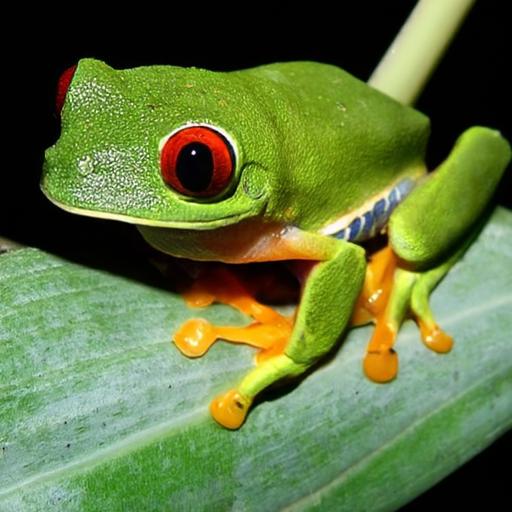} \hfill
    \includegraphics[width=.12\textwidth]{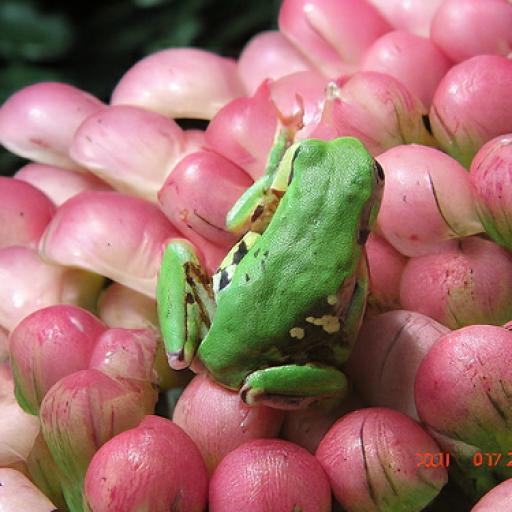} \hfill
    \includegraphics[width=.12\textwidth]{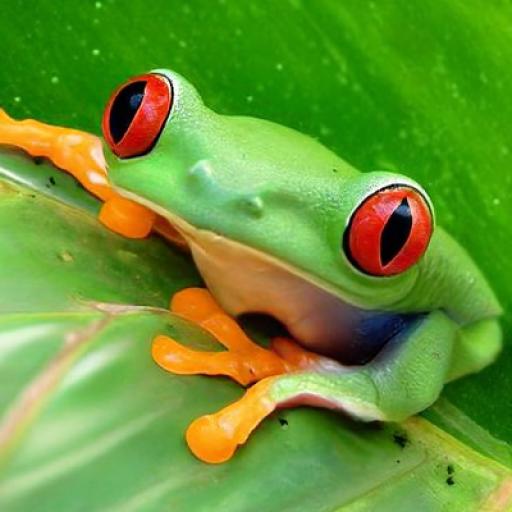} \hfill
    \includegraphics[width=.12\textwidth]{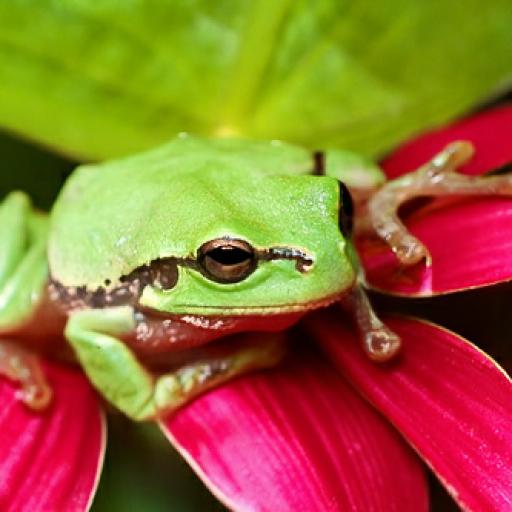} \hfill
    \includegraphics[width=.12\textwidth]{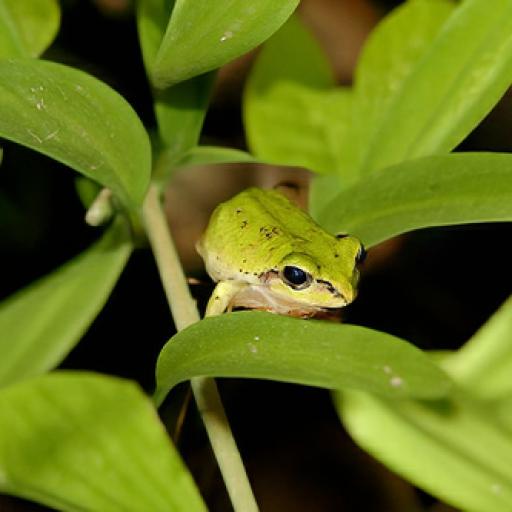} \hfill
    \includegraphics[width=.12\textwidth]{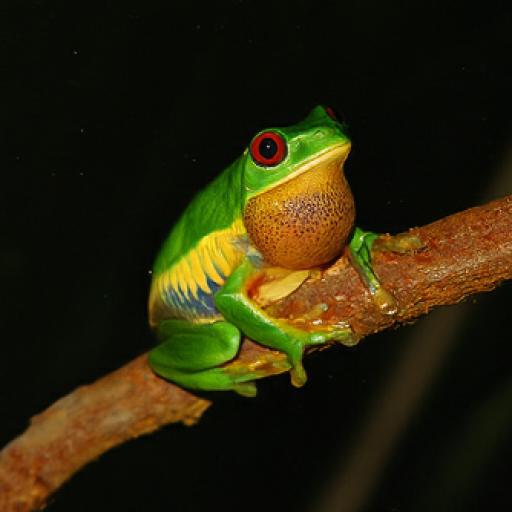} \hfill
    \includegraphics[width=.12\textwidth]{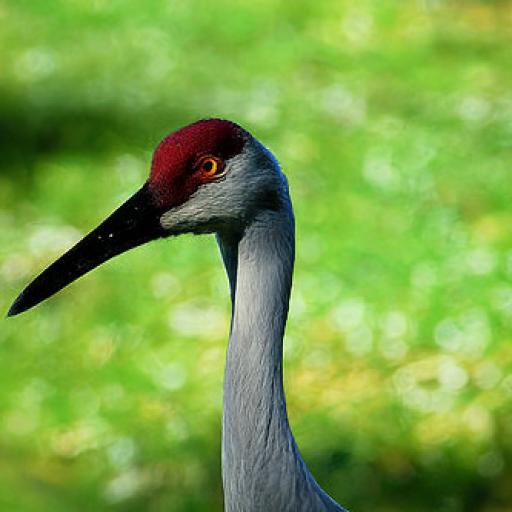} \hfill
    \includegraphics[width=.12\textwidth]{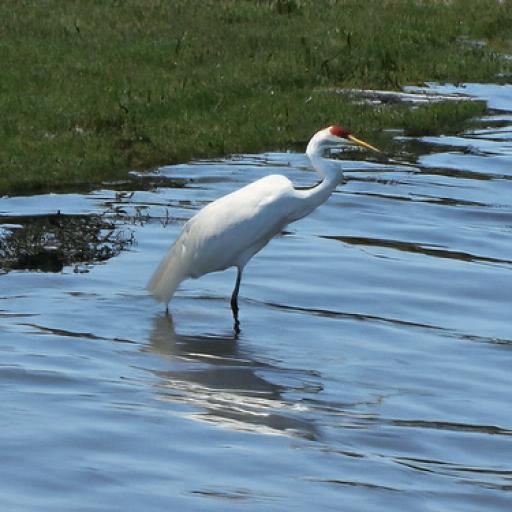} \hfill
    \includegraphics[width=.12\textwidth]{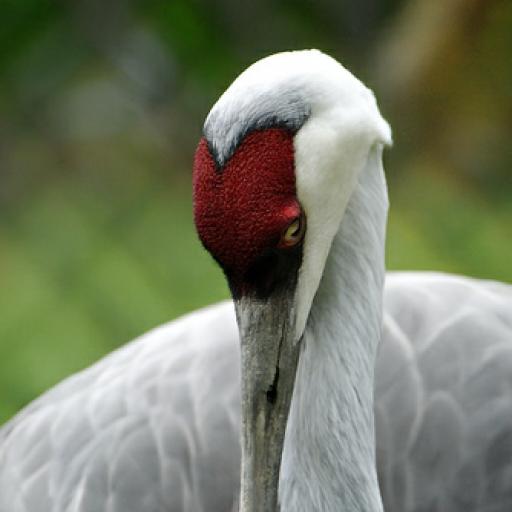} \hfill
    \includegraphics[width=.12\textwidth]{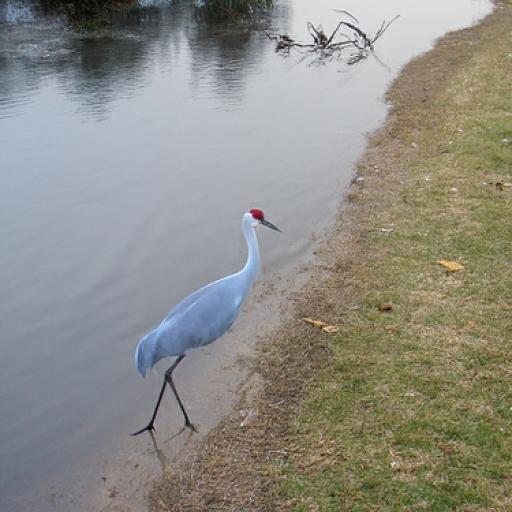} \hfill
    \includegraphics[width=.12\textwidth]{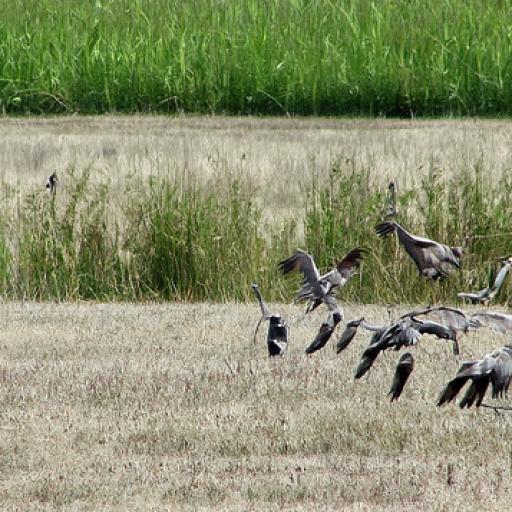} \hfill
    \includegraphics[width=.12\textwidth]{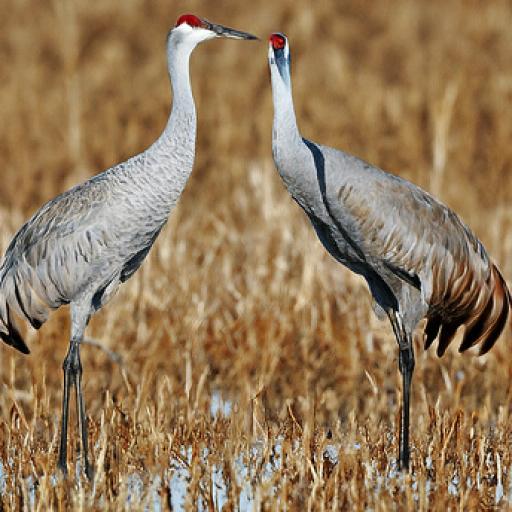} \hfill
    \includegraphics[width=.12\textwidth]{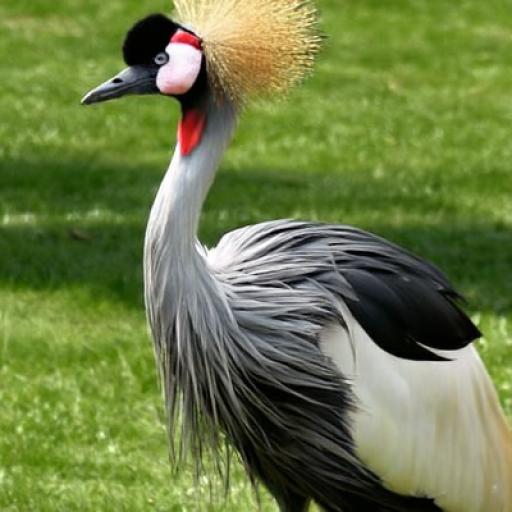} \hfill
    \includegraphics[width=.12\textwidth]{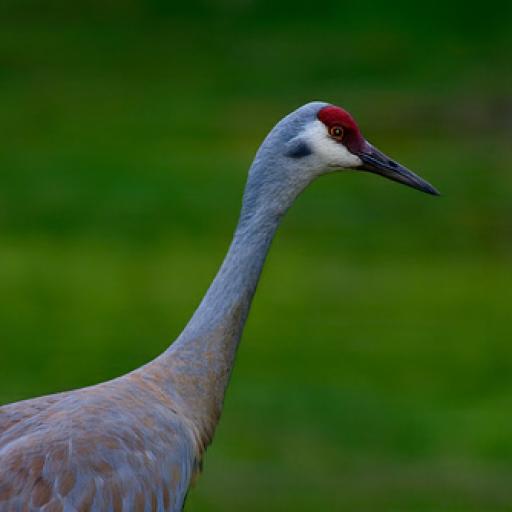} \hfill
    \includegraphics[width=.12\textwidth]{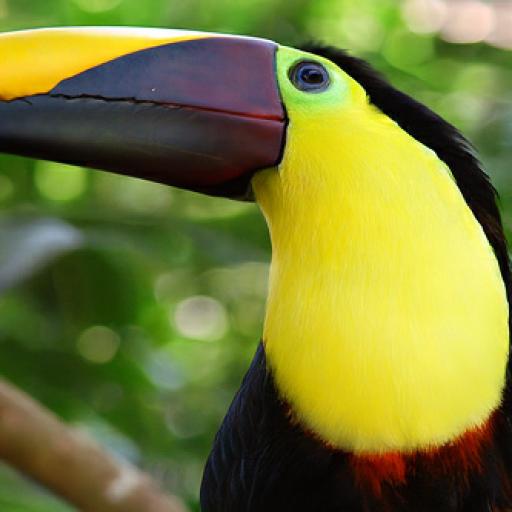} \hfill
    \includegraphics[width=.12\textwidth]{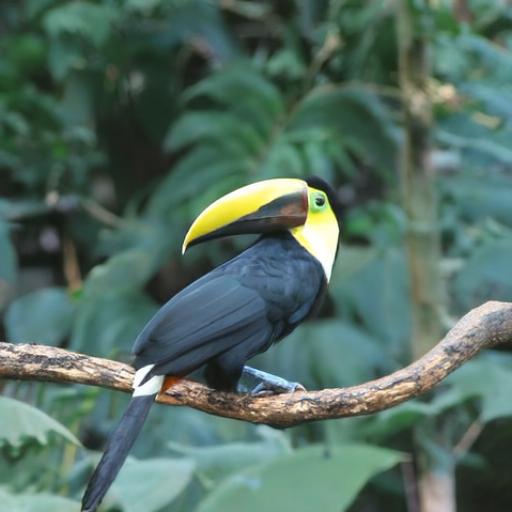} \hfill
    \includegraphics[width=.12\textwidth]{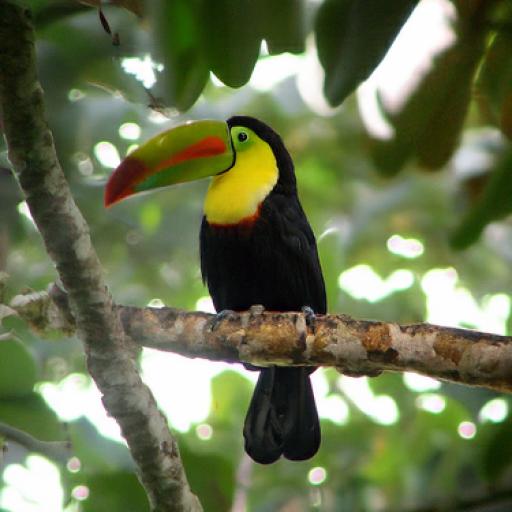} \hfill
    \includegraphics[width=.12\textwidth]{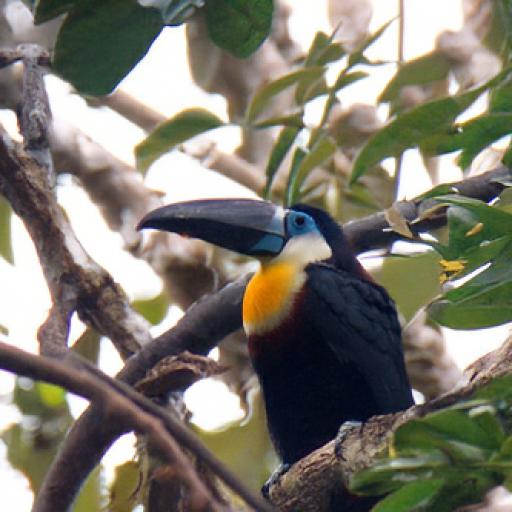} \hfill
    \includegraphics[width=.12\textwidth]{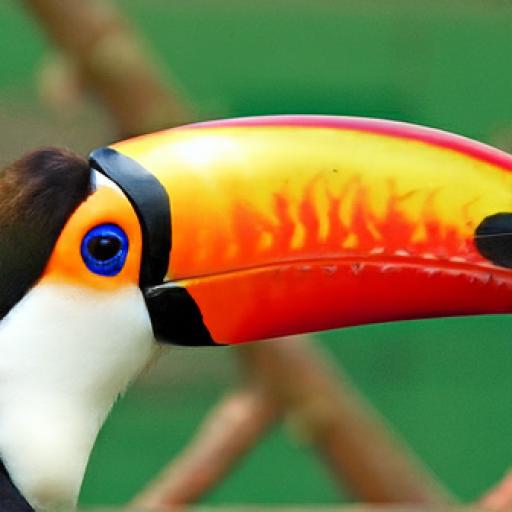} \hfill
    \includegraphics[width=.12\textwidth]{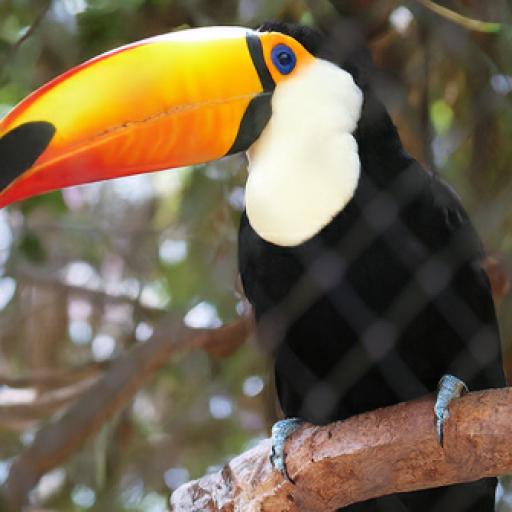} \hfill
    \includegraphics[width=.12\textwidth]{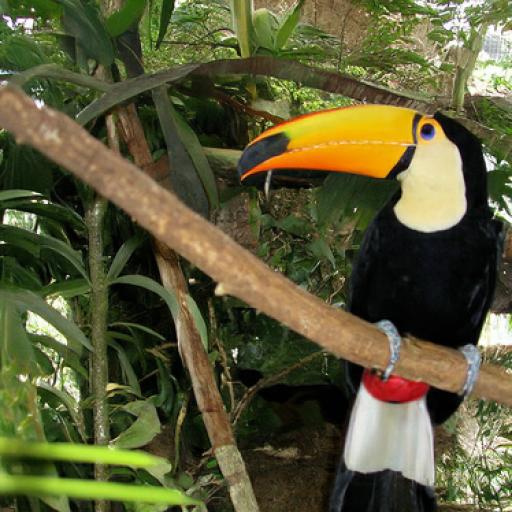} \hfill
    \includegraphics[width=.12\textwidth]{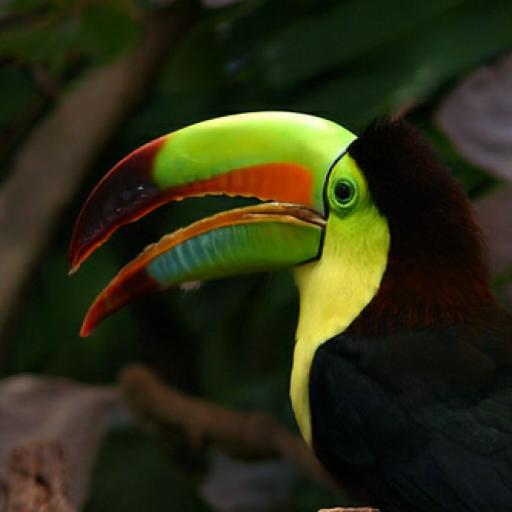}
    \caption{Some \textit{random (\textbf{not} cherry-picked)} class-conditional samples of $512 \times 512$ that have been generated by SiD2 flop heavy. Guidance 2.0 on interval logsnr $\in$ (-8, 5). Every row uses the same class-conditioning, every column uses the same rng.
    }
    \label{fig:class_conditional_imagenet}
\end{figure*}

\begin{figure*}
    \centering
    \resizebox{.99\textwidth}{!}{
    \parbox{.48\textwidth}{
        \includegraphics[width=.11\textwidth]{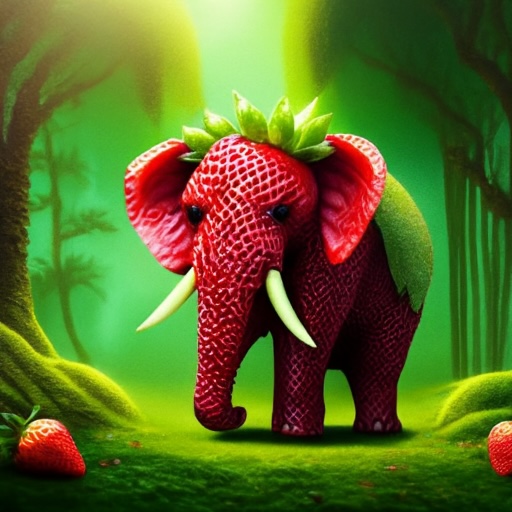} 
        \includegraphics[width=.11\textwidth]{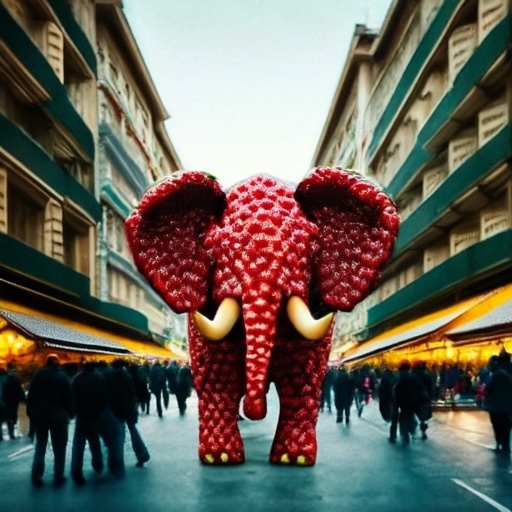} \quad
        \includegraphics[width=.11\textwidth]{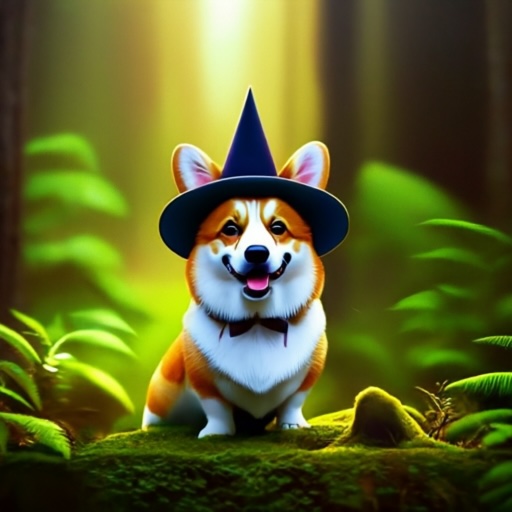} 
        \includegraphics[width=.11\textwidth]{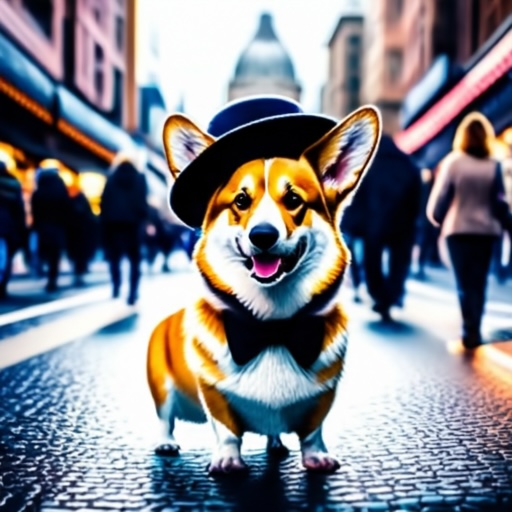} \\
        \includegraphics[width=.11\textwidth]{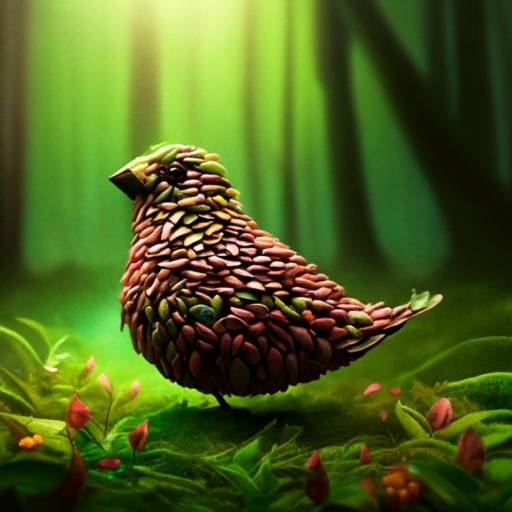}
        \includegraphics[width=.11\textwidth]{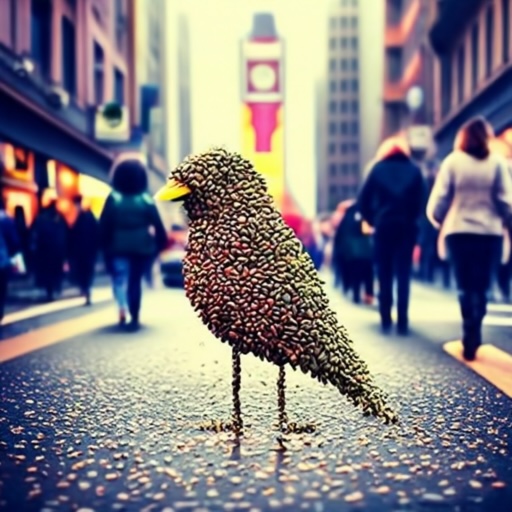} \quad
        \includegraphics[width=.11\textwidth]{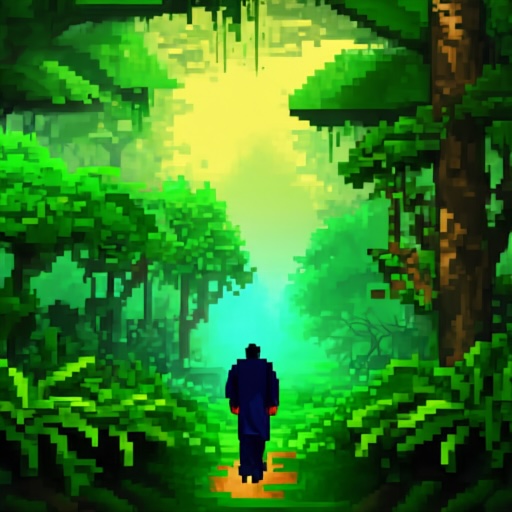}
        \includegraphics[width=.11\textwidth]{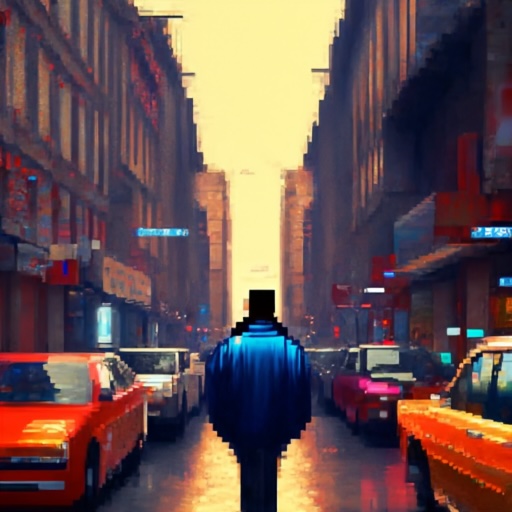}
        }
    }
    \caption{Generations from the text-to-image model. \textbf{Used prompts} (the Cartesian product of the sets): "\textit{\{A strawberry elephant, A corgi wearing a hat, A bird made of seeds, Pixel art: a man walking\}} in a \textit{\{lush forest, busy street\}.}"
    }
    \label{fig:text_to_image}
\end{figure*}

\begin{figure*}
    \centering
    \resizebox{.99\textwidth}{!}{
    \parbox{.468\textwidth}{
        \includegraphics[width=.11\textwidth]{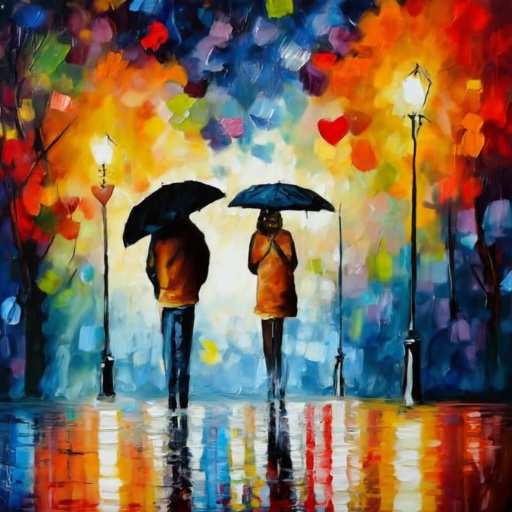}
        \includegraphics[width=.11\textwidth]{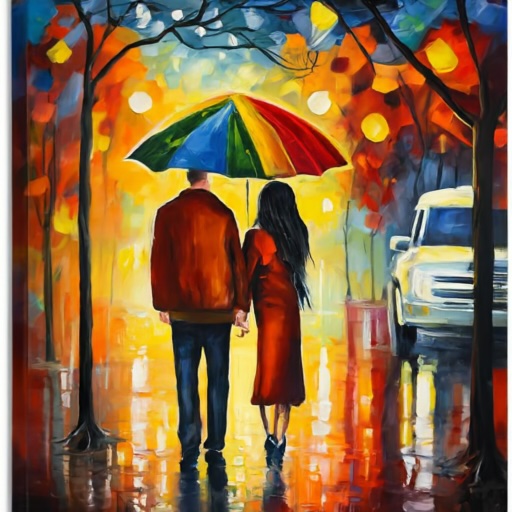}\quad
        \includegraphics[width=.11\textwidth]{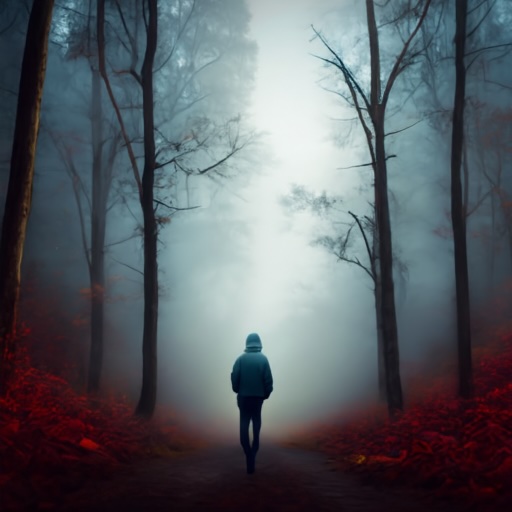} 
        \includegraphics[width=.11\textwidth]{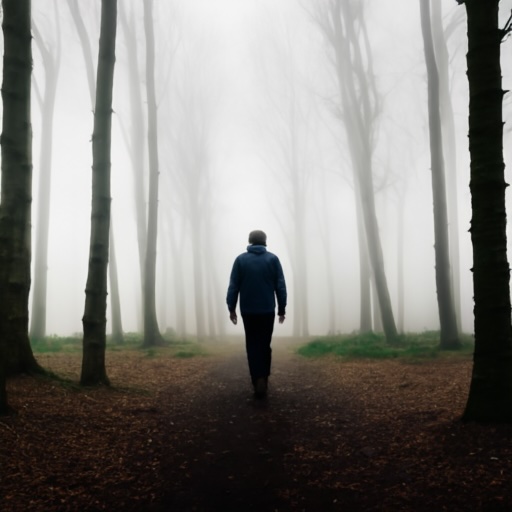} \\
        \includegraphics[width=.11\textwidth]{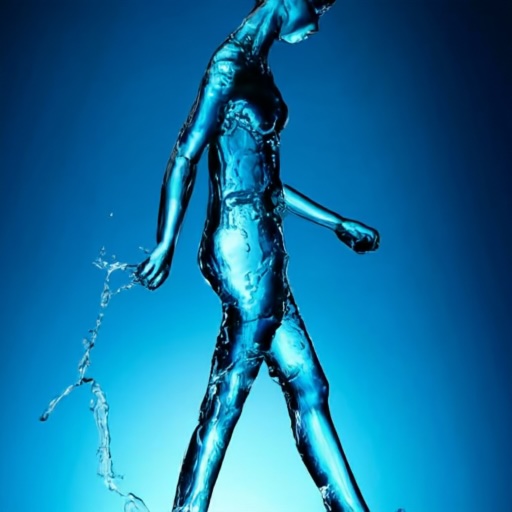}
        \includegraphics[width=.11\textwidth]{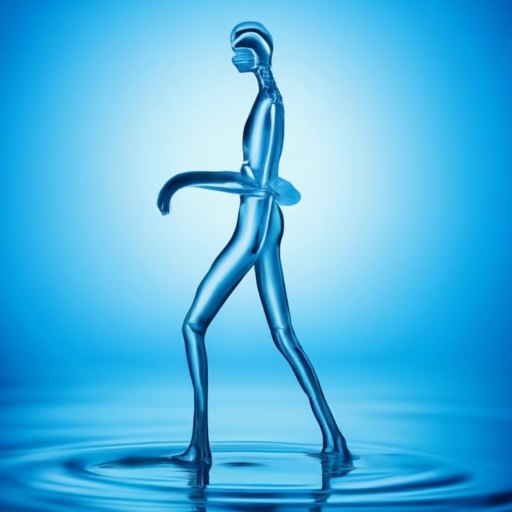}\quad
        \includegraphics[width=.11\textwidth]{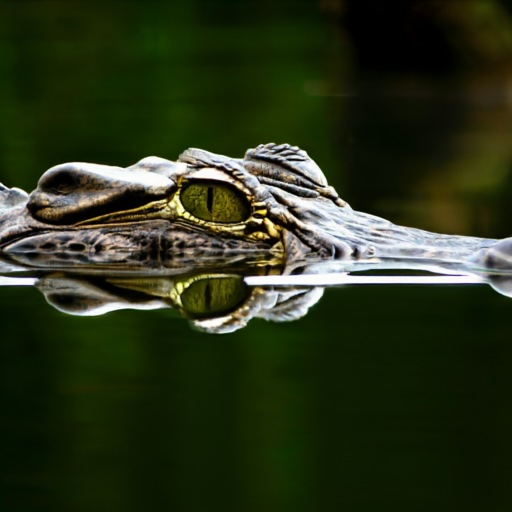} 
        \includegraphics[width=.11\textwidth]{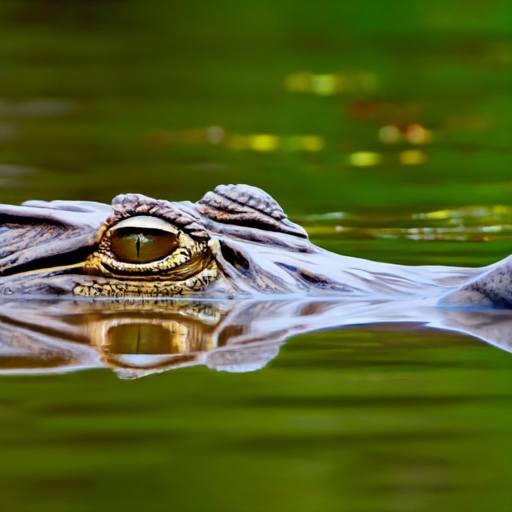} \\
        \includegraphics[width=.11\textwidth]{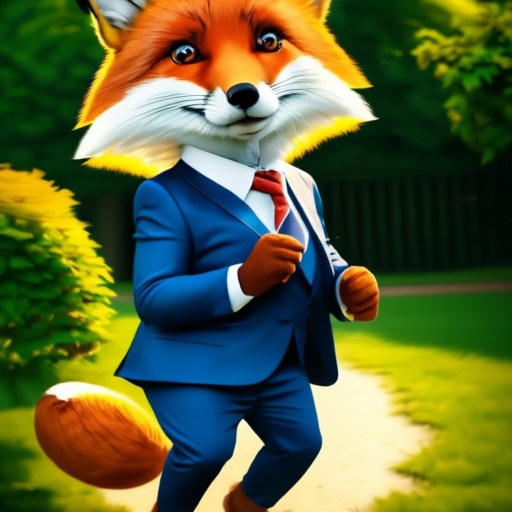}
        \includegraphics[width=.11\textwidth]{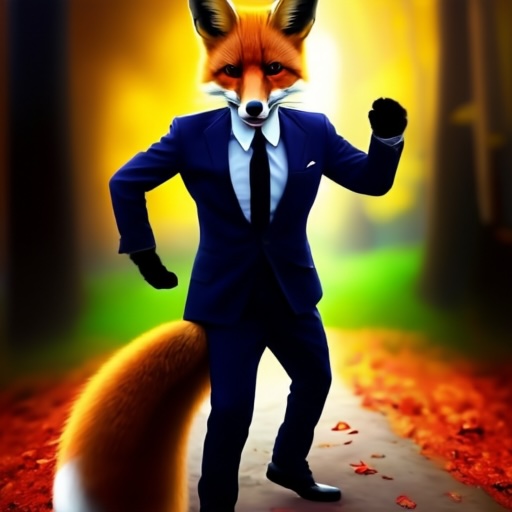}\quad
        \includegraphics[width=.11\textwidth]{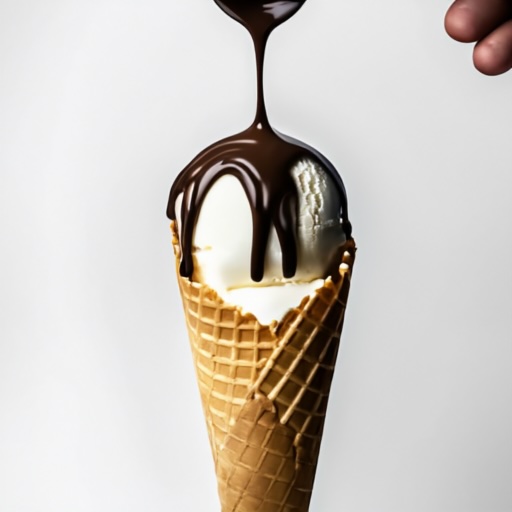} 
        \includegraphics[width=.11\textwidth]{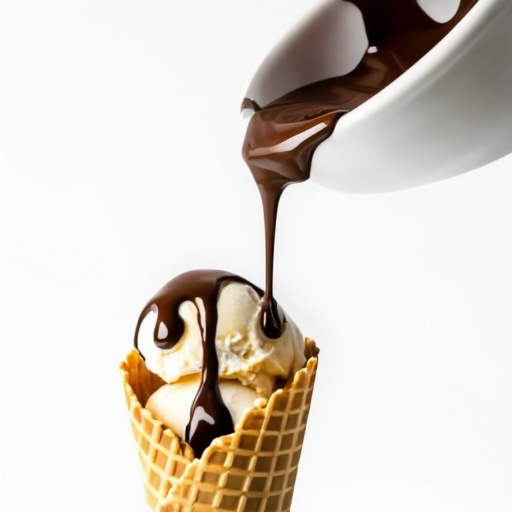} \\
        \includegraphics[width=.11\textwidth]{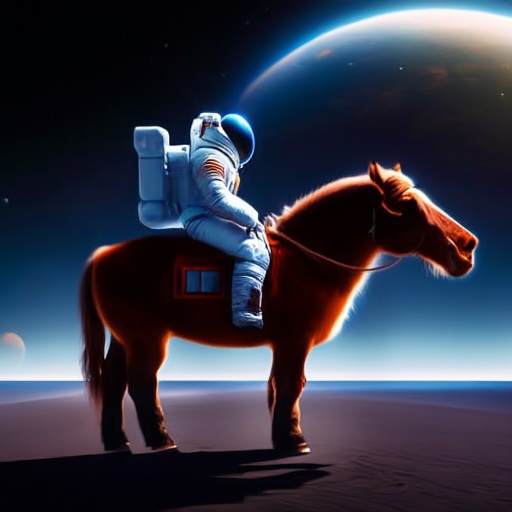}
        \includegraphics[width=.11\textwidth]{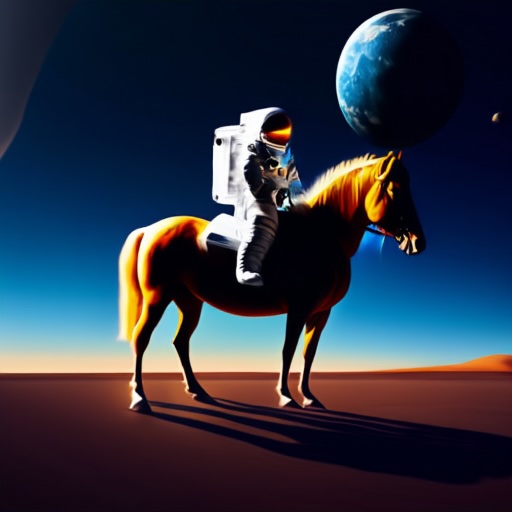}\quad
        \includegraphics[width=.11\textwidth]{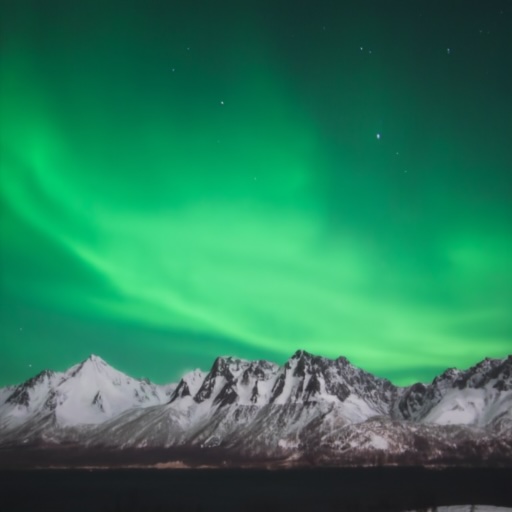} 
        \includegraphics[width=.11\textwidth]{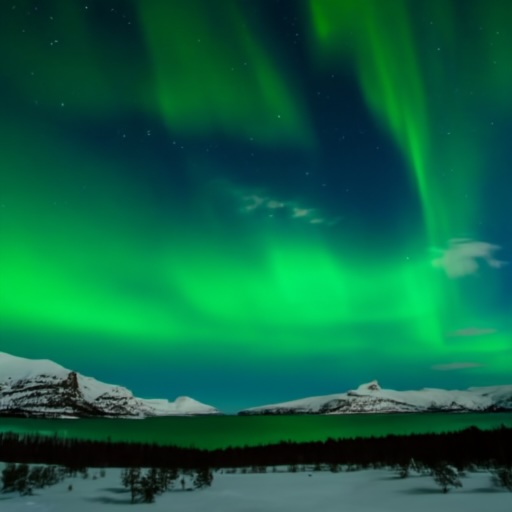} \\
        \includegraphics[width=.11\textwidth]{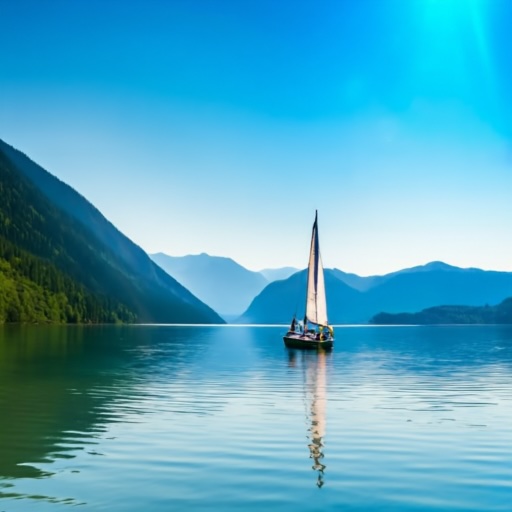}
        \includegraphics[width=.11\textwidth]{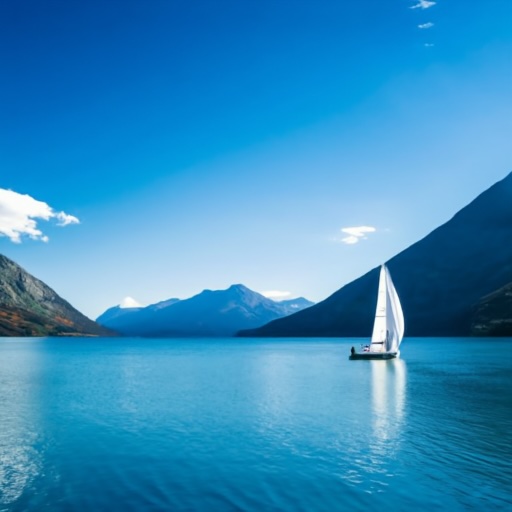}\quad
        \includegraphics[width=.11\textwidth]{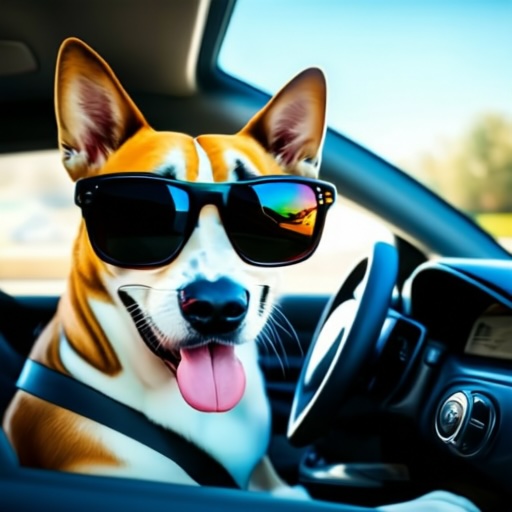} 
        \includegraphics[width=.11\textwidth]{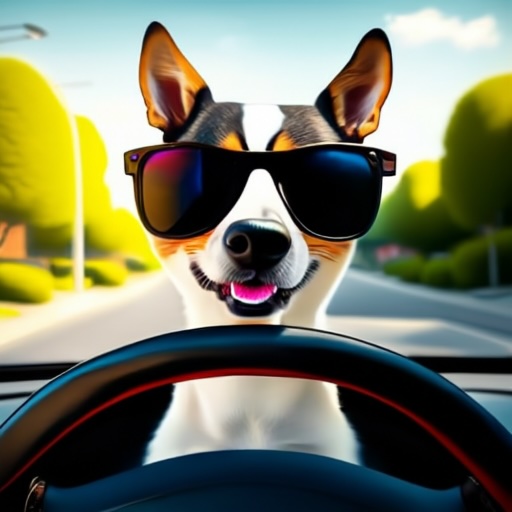} 
        }
    }
    \caption{Generations from the distilled text-to-image model. Guidance is set to +3. For each prompt two images selected from four generated images. \textbf{Used prompts}:
(1) \textit{A couple gets caught in the rain, oil on canvas},
(2) \textit{A lone traveller walks in a misty forest},
(3) \textit{A walking figure made out of water},
(4) \textit{In the swamp, a crocodile stealthily surfaces, revealing only its eyes and the tip of its nose as it moves forward},
(5) \textit{A fox dressed in suit dancing in park},
(6) \textit{Pouring chocolate sauce over vanilla ice cream in a cone, studio lighting},
(7) \textit{An astronaut riding a horse},
(8) \textit{Aurora Borealis Green Loop Winter Mountain Ridges Northern Lights},
(9) \textit{Sailboat sailing on a sunny day in a mountain lake},
(10) \textit{A dog driving a car on a suburban street wearing funny sunglasses}.
    }
    \label{fig:text_to_image_2}
\end{figure*}

We provide more illustrative examples for ImageNet512 generation (Figure~\ref{fig:class_conditional_imagenet}) and Text-to-Image (Figure~\ref{fig:text_to_image} and Figure~\ref{fig:text_to_image_2}).

\newpage

\end{document}